\documentclass[preprint,1p,times]{elsarticle}
\usepackage{amssymb}
\usepackage{amsmath}
\usepackage{multirow}
\usepackage{graphicx}
\usepackage{amsmath,amsfonts}
\usepackage{algorithmic}
\usepackage{algorithm}
\usepackage{array}
\usepackage[caption=false,font=normalsize,labelfont=sf,textfont=sf]{subfig}
\usepackage[table,xcdraw]{xcolor}
\usepackage{textcomp}
\usepackage{mathtools}
\usepackage{stfloats}
\usepackage{url}
\usepackage{float}
\usepackage{verbatim}
\usepackage{graphicx}
\usepackage{subfig}
\usepackage{tabto}
\usepackage{multirow}
\usepackage{hhline}
\usepackage{amsmath}
\usepackage{hyperref}
\usepackage{color,soul}
\usepackage{eqnarray,amsmath}
\usepackage{geometry}
\sethlcolor{white}
\usepackage[cal=cm]{mathalpha}
\journal{}

\begin{document}

\begin{frontmatter}



\title{Industrial-scale Prediction of Cement Clinker Phases using Machine Learning}


\author[1]{Sheikh Junaid Fayaz} %% Author name
\author[2]{Néstor Montiel-Bohórquez} %% Author name
\author[1]{Shashank Bishnoi} %% Author name
\author[2]{Matteo Romano} %% Author name
\author[2]{Manuele Gatti} %% Author name
\author[1,*]{N. M. Anoop Krishnan} %% Author name

\affiliation[1]{organization={Indian Institute of Technology Delhi},%Department and Organization
            city={Hauz Khas},
            postcode={110016}, 
            state={New Delhi},
            country={India}}

\affiliation[2]{organization={Politecnico di Milano, Department of Energy},%Department and Organization
            addressline={Lambruschini 4A, 20156}, 
            city={Milan},
            country={Italy}}
\affiliation[*]{Corresponding authors: NMAK (krishnan@iitd.ac.in)
}
%% Abstract

\begin{abstract}

Cement production, exceeding 4.1 billion tonnes and contributing 2.4 tonnes of CO$_2$ annually, faces challenges in quality control and process optimization. Traditional process models for cement manufacturing are confined to steady-state conditions with limited predictive capability for mineralogical phases. However, modern plants operate dynamically, requiring real-time quality assessment. Here, exploiting a comprehensive two-year operational dataset from an industrial cement plant, we present a machine learning framework that accurately predicts clinker mineralogy from process data. Our model achieves unprecedented accuracy for predicting major clinker phases while requiring minimal inputs, demonstrating robust performance under varying operations. Through post-hoc explainable algorithms, we interpret the hierarchical relationships between clinker oxides and phase formation, providing insights into the functioning of an otherwise black-box model. This digital twin framework can potentially enable real-time optimization of cement production, thereby providing a route toward reducing material waste and ensuring quality while reducing the associated emissions under real-world conditions. 
% Our approach represents a significant advancement in industrial process control, offering a scalable solution for sustainable cement manufacturing.

%Finally, we develop a set of first-order linear equations with a similar form to Bogue’s formula, which substantially improves prediction accuracy, offering a more reliable alternative for the industry.
\end{abstract}

%% Keywords
\end{frontmatter}

%% the following line to enable line numbers



\section{Introduction}
% Cement is the most produced human-made material in the world \cite{schneider_sustainable_2011}. 
Global cement production reached more than $4.1$ billion tons/year in $2023$, more than doubling from 2005 levels~\cite{kim_slag_2018,panesar_effect_2019}, with projections indicating a further increase by 2050~\cite{panesar_effect_2019}. Moreover, each tonne of cement releases $\sim$0.6 tonne CO$_2$, contributing to more than 8\% of global carbon emissions. This unprecedented growth~\cite{cancio_diaz_limestone_2017}, along with high emissions, necessitates enhanced production efficiency while maintaining stringent quality standards. The cement's performance characteristics, particularly its 28-day compressive strength, are fundamentally determined by the clinker's mineralogical phases---alite, belite, aluminate, and ferrite. Determining the mineralogical phases directly from the clinker plays a crucial role in determining the quality of the cement produced at the plant~\cite{zaki_cementron_2023}.

Quality assessment of clinker mineralogy traditionally relies on X-ray diffraction (XRD), performed either online with 15-30 minute delays or offline with up to 4-hour measurement cycles~\cite{negash_neural_2012}. These delays result in significant material waste when out-of-specification clinker is produced. Real-time prediction of clinker phases would not only eliminate this waste but also enable proactive process control through feed composition and operating parameter adjustments.
While process modeling has successfully addressed various aspects of cement production, including calciner operations \cite{sharma_aspen_2023}, \cite{zhang_aspen_2011}, waste heat recovery \cite{redjeba_aspenplus_2019}, \cite{rahman_aspen_2014}, alternative fuel assessment \cite{kaantee_cement_2004}, and CO$_{2}$ capture \cite{arachchige_model_2013}, accurate prediction of clinker mineralogy remains elusive. The complexity of multi-phase equilibria and thermochemical phenomena in rotary kilns presents formidable modeling challenges. Despite efforts to model kiln dynamics \cite{meyer2016computation}, reliable estimation of clinker mineralogy continues to be an open problem.

First principle models (FPMs) \cite{mastorakos_cfd_1999,wang_dynamic_2008,darabi_mathematical_2007,sadighi_rotary_2011} have attempted to address computational efficiency through steady-state approximations \cite{moses_predictive_2016}. These models, for example, comprising twelve differential-algebraic and two algebraic equations based on mass balance, energy balance, and material residence time, showed promising results for alite prediction using response surface modeling \cite{dean_response_2017}. Alternatively, recent advances in machine learning (ML) in materials domain~\cite{krishnan2024machine} demonstrate promising capabilities in process modeling~\cite{li_machine_2022,ali_machine_2022,jablonka_machine_2023} and predicting cement and concrete properties~\cite{lyngdoh2022prediction,lapeyre2021machine,miyan2024integrating}. However, these approaches have been primarily validated on limited laboratory-scale data, leaving their efficacy for continuous industrial operations and complex plant-wide dynamics largely unexplored. A fundamental question remains: can ML potentially transform cement manufacturing by creating accurate digital twins from industrial-scale data, thereby enabling real-time process optimization?

To address this challenge, we leverage a comprehensive two-year industrial dataset from an operating cement plant to develop predictive models for clinker mineralogical phases. Our framework demonstrates unprecedented accuracy in phase composition prediction, significantly outperforming traditional Bogue equations, offering a practical pathway toward automated process control. Through post-hoc explainability methods, we elucidate the quantitative relationships between clinker oxides and phase formation dynamics. Finally, we establish plant-specific equations that substantially exceed conventional Bogue calculations in prediction accuracy, providing a quick assessment tool for plant-scale operations.

\section{Methods}
\subsection{Data collection}

The operational dataset spans from 01/01/2020 to 31/12/2021, collected from an industrial cement plant facilitated through the Innovandi consortium. The database architecture comprised three distinct components:
\begin{itemize}
\item \textbf{DB0}: Plant configuration parameters including kiln specifications, pre-calciner characteristics, preheater architecture (strings and stages), and bypass systems.
\item \textbf{DB1}: Process parameters including stage-wise temperature and pressure profiles, O$_2$ content, kiln feed temperature, and calciner fuel consumption rates. Complete specifications are provided in Table~\ref{Table_A1}, ref{appendix a data} .
\item \textbf{DB2}: Compositional analyses of kiln feed (KF), hot meal (HM), and clinker, including oxide distributions and phase compositions (Table~\ref{Table_A2}, \ref{appendix a data}). All compositions are reported as weight percentages (wt.\%).
\end{itemize}
Figure~\ref{fig1new}(a) illustrates the plant schematic and data collection points for PP, KF, HM and clinker composition. Mineral oxide compositions in KF, HM, and clinker were quantified using X-ray fluorescence (XRF) spectroscopy. The percentage of clinker phases was determined through X-ray diffraction (XRD) analysis. 

Data synchronizing by identifying the process timeline is a crucial aspect to meaningfully identify the relevant input parameters corresponding to a given output clinker composition. The clinker production follows a sequential timeline:
\begin{enumerate}
\item KF measurement at preheater tower inlet ($t_0$)
\item One-minute buffer period preceding kiln feed ($1$ minute; total time: ($t_0+1$) minutes)
\item Preheater tower residence ($\sim16$ minutes; total time: ($t_0+17$) minutes)
\item Clinker cooler retention ($\sim20$ minutes; total time: ($t_0+37$) minutes)
\item Post-production sampling delay ($\sim20$ minutes; total time: ($t_0+57$) minutes)
\end{enumerate}
The total duration from feed introduction to clinker formation approximates 37 minutes. The clinker composition data was timestamped based on production time, while KF, HM, and process parameter data were timestamped according to measurement time, enabling precise temporal correlation between input parameters and output characteristics.

\subsection{Data processing}
The two-year operational dataset presented unique challenges characteristic of industrial-scale data collection. Measurement uncertainties stemming from instrumental limitations and human factors necessitated rigorous preprocessing to ensure data integrity. The raw dataset exhibited three primary irregularities: duplicate entries, missing measurements, and physically inconsistent values (e.g., non-normalized XRD measurements, out-of-range variables, XRF-XRD mismatches).
We implemented a systematic three-tier preprocessing protocol:
\begin{enumerate}
\item Data Completeness: Initial screening eliminated 207 clinker compositions lacking corresponding KF, HM, and process parameter measurements.
\item Data Consistency: 
\begin{itemize}
    \item Consolidated duplicate entries
    \item Removed incomplete feature sets
    \item Applied 0.01-99.99 percentile filtering to exclude non-representative outliers and plant shutdown periods
\end{itemize}
\item Physical Validation: Enforced compositional constraints to ensure thermodynamic consistency of phase measurements.\end{enumerate}
The outlier threshold selection balanced data retention with statistical significance. As evidenced in Fig.~\ref{fig1new} b-d, excluded data points ($<50$ per variable) deviated significantly from the two-year compositional distributions. This filtering methodology was consistently applied across all variables to prevent spurious correlations during model training.
The final curated dataset comprised 8,654 clinker measurements, representing approximately 58\% of the raw data. Detailed preprocessing statistics are provided in Table~\ref{Table_A3},~\ref{appendix a data}.

\subsection{ML models}
Four primary feature categories - process parameters (PP), kiln feed (KF), hot meal (HM), and clinker oxides (CO) - generated 15 unique input combinations (Fig.~\ref{fig1new}a). Each combination underwent evaluation through multiple algorithmic architectures: linear regression, lasso, elastic net, support vector regression, random forest, XGBoost, neural networks, and Gaussian processes. Mathematical formulations of these modes are provided in \ref{ML models}~\cite{krishnan2024machine}. All the models were trained using the sklearn package.

\subsection{Performance Metrics}
Model assessment utilized three complementary metrics described below.

(i) Mean Absolute Percentage Error (MAPE). \cite{montano_moreno_using_2013}:
\begin{equation}
MAPE=\frac{1}{n}  \sum_{i=1}^{n} \frac{|y_p\ (i)-y_t\ (i)|}{y_t\ (i)}
\end{equation}
where $n$ represents sample size, $y_p(i)$ and $y_t(i)$ denote predicted and true values.

(ii) Mean Absolute Error (MAE):
\begin{equation}
MAE=\frac{1}{n}  \sum_{i=1}^{n} |y_p\ (i)-y_t\ (i)|
\end{equation}

(iii) Coefficient of Determination $(R^{2})$ \cite{noauthor_pearson_nodate}:
\begin{equation}
R^2= 1-\frac{\textnormal{RSS}}{\textnormal{TSS}}\
\end{equation}
where,
\begin{align*}
\textnormal{RSS}= \sum_{i=1}^{n} (y_p(i)-y_t(i))^2 &&
\textnormal{TSS}= \sum_{i=1}^{n} (y_p(i)-\Bar{y})^2 &&
\bar{y}=\frac{1}{n} \sum_{i=1}^{n} (y_p(i)-y_t(i))^2
\end{align*}

\subsection{Model Optimization Protocol}
Hyperparameter optimization followed a systematic protocol~\cite{krishnan2024machine}:
\begin{enumerate}
    \item Data partitioning: 80:20 train-test split maintaining statistical equivalence via sklearn~\cite{noauthor_sklearnpreprocessingstandardscaler_nodate}. 
    \item Parameter optimization: 4-fold cross-validation \cite{bates_cross-validation_2023} utilizing GridSearchCV~\cite{noauthor_sklearnmodel_selectiongridsearchcv_nodate}.
    \item Model selection: Optimization based on validation scores (Table~\ref{Table_supp}, \ref{appendix model training}).
    \item Validation: Final assessment on holdout test set.
\end{enumerate}
Detailed optimization protocols appear in \ref{appendix model training}. The hyperparametric optimization was performed using the GridSearchCV in the sklearn package.

\subsection{Shapley Additive Explanations (SHAP)}
To interpret the learned black-box functions within complex ML architectures, we implemented SHAP analysis \cite{lundberg_unified_2017}, a post-hoc interpretation framework derived from game-theoretic Shapley values \cite{shapley_value_1953}. This methodology quantifies individual feature contributions to model predictions through systematic feature importance allocation.
The SHAP value ($\chi_k$) for feature $k$ represents its average marginal contribution across all possible feature subsets:
\begin{equation}
\chi(f,X)= \sum_{\ddot{z} \subset X'} \frac{\lvert \ddot{z}\rvert (n-\lvert \ddot{z}\rvert -1)!}{n!}f_X-f_X(\ddot{z}_{\backslash k})]
\end{equation}
where, $f$: Original model architecture, $X$: Complete feature set, $\lvert \ddot{z}\rvert$: Cardinality of feature combinations within non-zero feature power set, $n$: Simplified input feature count, $\ddot{z} \subset X^{\prime}$: Vectors with non-zero entries subset to $\ddot{X}$, $f_X(\ddot{z})$: Model prediction for given data point, $f_X(\ddot{z}_{\backslash k})$: Model prediction excluding feature $k$ \cite{waris2024pseudo}

SHAP values correlate directly with prediction error magnitude - larger errors indicate greater feature importance. The framework accommodates various model-specific approximations (kernel, Deep, Linear, Tree-based explainers), reducing complex architectures to tractable polynomial forms \cite{bhattoo_understanding_2023,krishnan2024machine}. Comprehensive theoretical foundations can be found in \cite{lundberg_unified_2017}. SHAP library was employed to perform the SHAP computations.

\section{Results} 
\subsection{Cement plant data}
\label{dataset visualization}

% \newgeometry{bottom=2.5cm}
\begin{figure}[hbtp]
    \centering
    \includegraphics[scale=0.6]{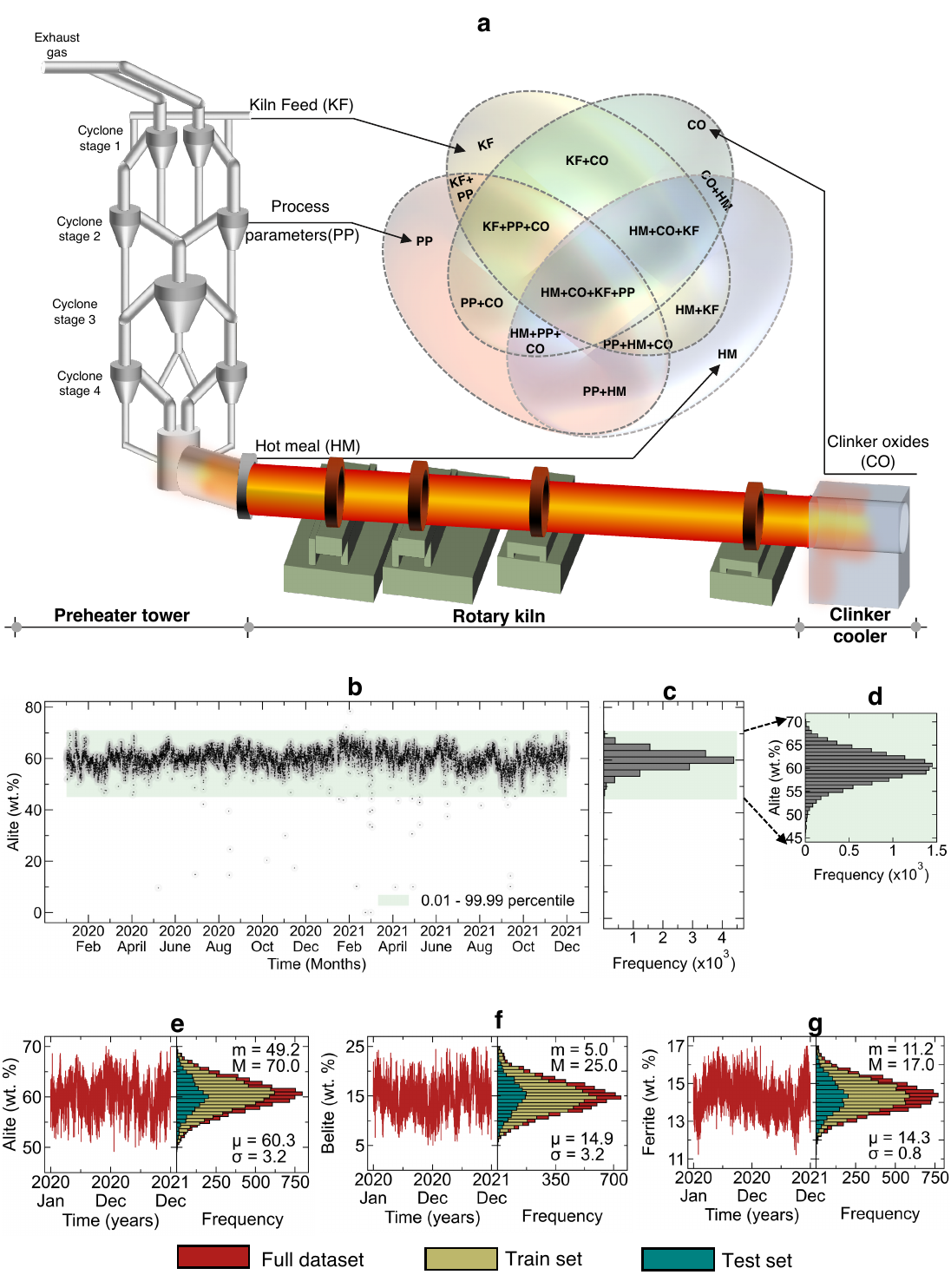}
    \caption{\textbf{Dataset characteristics and temporal variability in clinker phases.} a, Schematic representation of a cement plant showing key measurement locations: kiln feed (KF), process parameters (PP), hot meal (HM), and clinker oxides (CO). The Venn diagram illustrates the combinations of input features used for model development. b, Two-year temporal evolution of alite content showing plant variability (black dots) with 0.01-99.99 percentile bounds (green shading). c,d, Frequency distribution of alite content in the complete dataset and zoomed view highlighting the normal distribution. e-g, Time series and distribution analysis of major clinker phases (alite, belite, ferrite) showing data partitioning into training (yellow) and test (green) sets. Statistical parameters ($\mu$, $\sigma$) characterize the phase distributions. Each subplot includes temporal evolution (left) and frequency distribution (right) with mean (m) and standard deviation values. The full dataset is shown in red, the training set in yellow, and the test set in green.}
    \label{fig1new}
    \end{figure}

A comprehensive two-year (01/01/2020 to 31/12/2021) operational dataset from an industrial cement plant formed the foundation for our ML framework. To the best of our knowledge, this is the first time that such a large-scale data from the cement plant has been used to develop a data-driven digital twin. The data architecture comprised three distinct databases: plant configuration parameters, namely, DB0; continuous process parameters (PP), namely, DB1; and compositional analyses, namely, DB2, including fuel characteristics, kiln feed (KF), hot meal (HM), and clinker compositions including clinker oxides (CO). The material compositions were determined through XRD and XRF analyses. PP comprised on 34 features, while KF, HM, and clinker comprised of 9, 7, and 12 features, respectively. Thus, the total dataset consists of potentially 59 input features with the three clinker phases as the output. Complete details of the features in DB1 and DB2 are reported in Tables~\ref{Table_A1} and ~\ref{Table_A2} in~\ref{appendix a data}.

The temporal resolution varied significantly across parameters (Table~\ref{Table_1}), necessitating careful data preprocessing. Process parameters exhibited high-frequency sampling (1-minute intervals), while material compositions were measured at lower frequencies: hourly for clinker and kiln feed, bi-hourly for hot meal (detailed distributions in Appendix~\ref{appendix a data}, Figs.~\ref{dist_PP_1}-\ref{hrm dist}). This multi-scale temporal architecture, combined with material transport dynamics through the kiln system, demanded a sophisticated synchronization framework to establish causal relationships between process conditions and resultant clinker compositions.

To address this challenge, we implemented a systematic temporal synchronization protocol to account for varying sampling frequencies and material transport dynamics. Each process measurement was standardized to 2-hour intervals through weighted temporal averaging, with specific consideration for the characteristic residence times: 1-minute buffer post-kiln feed measurement, 16-minute preheater tower residence, 20-minute clinker cooler retention, and 20-minute post-production sampling delay. The alignment algorithm correlated timestamps from process parameters (DB1) and compositional measurements (DB2) by incorporating these cumulative residence times ($\sim$37 minutes total), enabling precise mapping between input conditions and resulting clinker phases. This temporal registration ensured that each clinker composition measurement was accurately paired with its corresponding process conditions, accounting for material flow through the entire production chain from kiln feed to final clinker formation (see Methods for details).

Data quality assessment revealed significant variability in clinker phase compositions. An exemplar variation in alite content for the entire duration is shown in Fig.~\ref{fig1new}b, along with the respective distribution (Figs.~\ref{fig1new}c,d). The two-year alite measurements exhibited a broad distribution (45-70 wt.\%) with distinct temporal patterns. Similar variations were observed in belite (5-25 wt.\%) and ferrite (11-17 wt.\%) compositions (Fig.~\ref{fig1new}e-g). The complete distributions of all the features in DB1 and DB2 are included in Figs.~\ref{dist_PP_1}-\ref{hrm dist}. 

To ensure model robustness, we implemented a comprehensive outlier detection and data cleaning protocol. Values falling outside the 0.01-99.99 percentile range were classified as outliers and excluded from analysis, as illustrated by the shaded regions in Fig.~\ref{fig1new}b. This filtering criterion was uniformly applied across all 59 input features and 3 output variables. Following temporal synchronization, we performed completeness verification wherein any observation with missing values across any of the 62 features was excluded from the dataset. This rigorous preprocessing approach ensured a high-quality, complete dataset for subsequent analysis. The final dataset was split into training (70\%) and test (30\%) sets, maintaining the temporal distribution of phase compositions (Fig.~\ref{fig1new}e-g, shown in green and yellow respectively).

The preprocessed dataset exhibited normal distributions for all major phases, with alite centered at $\mu = 60.3$~wt.\% ($\sigma = 3.2$), belite at $\mu = 14.9$~wt.\% ($\sigma = 3.2$), and ferrite at $\mu = 14.3$~wt.\% ($\sigma = 0.8$). These distributions confirm that the data in the plant is indeed unbiased and is coming from a single distribution, confirming that operating conditions did not undergo any major change in the two-year period considered.  Thus, this data could be reliably used for developing predictive models while capturing the inherent variability in industrial operations.

\begin{table}[h!]
\caption{\label{Table_1}Details of the databases, DB1 and DB2, along with the data collection frequency, measurement techniques, and the total number of data points.}
\vspace{3mm}
\resizebox{\textwidth}{!}{%
\begin{tabular}{ccccc}
\hline 
\textbf{DB} &
  \textbf{Description} &
  \textbf{\begin{tabular}[c]{@{}c@{}}Data collection \\ frequency\end{tabular}} &
  \textbf{Measurement technique} &
  \textbf{\begin{tabular}[c]{@{}c@{}}Number of   \\ datapoints\end{tabular}} \\ \hline \\
DB1 & Process parameters & 1 minute         & Online measurement & 1,052,567 \\ \\
\multirow{3}{*}{DB2} &
  \multirow{3}{*}{\begin{tabular}[c]{@{}c@{}}Composition   \\ database\end{tabular}} &
  1 hour (for clinker) &
    &
  14,985 \\
    &                 & 1hour (for KF)   & XRD/XRF                  & 15,331    \\
    &                 & 2 hours (for HM) &            & 7,621     \\ \hline
\end{tabular}%
}
\end{table}

\subsection{Data-driven models for clinker phases}
\label{section ML performance}

\begin{figure}[hbtp]
    \centering
    \resizebox{0.8\textwidth}{!}{%
    \includegraphics[scale=1.0]{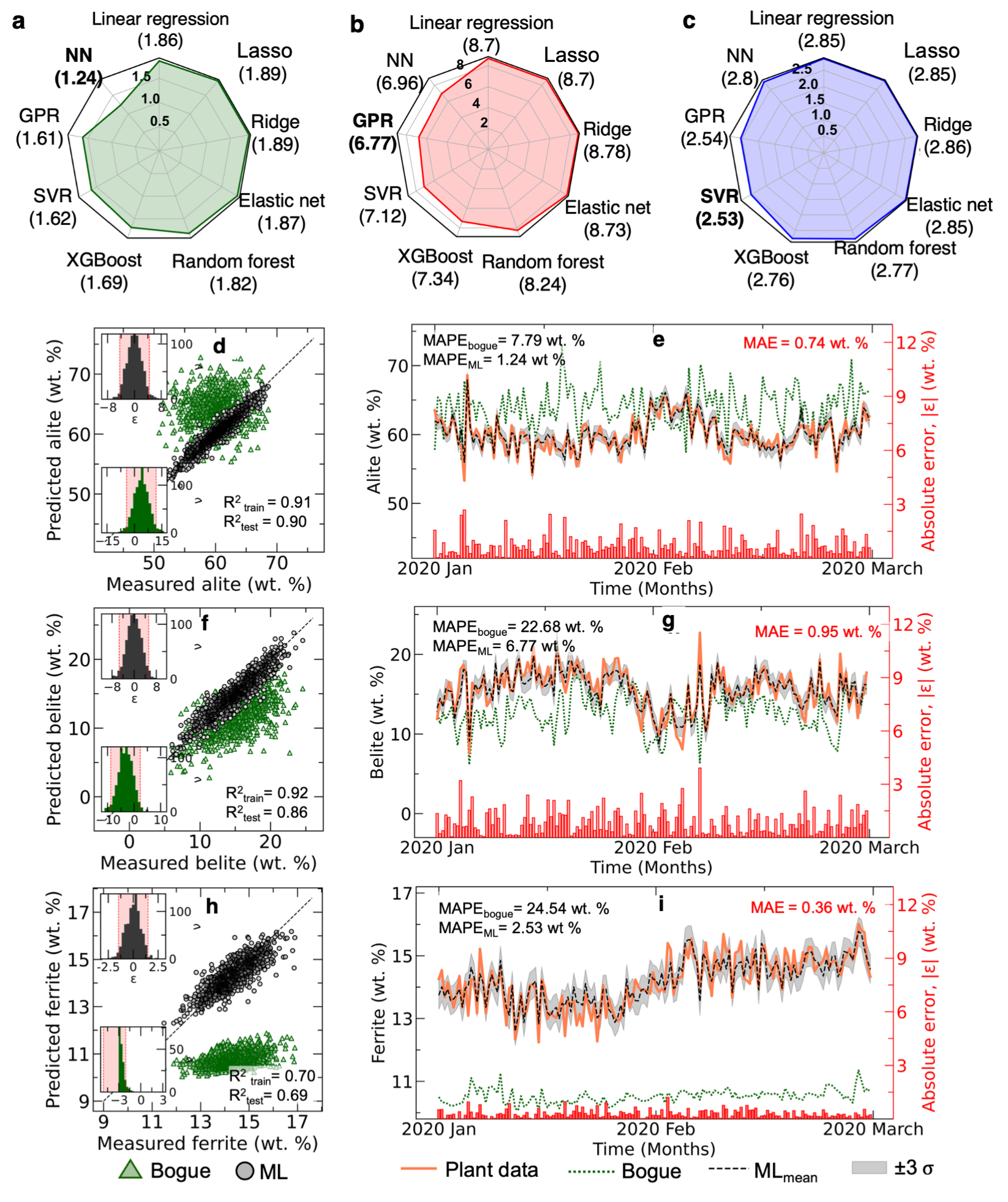}}
    \caption{\textbf{Performance comparison of machine learning architectures for clinker phase prediction.} a-c, Mean Absolute Percentage Error (MAPE) across nine ML models for predicting alite, belite, and ferrite compositions using complete feature sets (KF, PP, HM, CO), respectively. Values in parentheses indicate test set MAPE. The best-performing models are shown in bold. Quantitative performance metrics ($R^2$ and MAPE) for the best-performing model against traditional Bogue calculations represented as parity plot and temporal evolution, respectively, for d,e, alite, f,g, belite, and h,i, ferrite with inset histograms showing error distributions ($\epsilon = \text{predicted - actual}$) for ML models (top) and Bogue calculations (bottom). Red-shaded regions in histograms represent 95\% confidence intervals ($\pm 2\sigma$), with x-axis limits set at 99.9\% confidence ($\pm 4\sigma$). The temporal evolution of predictions is over a two-month test period showing plant data (red), ML predictions (black dashed), and Bogue calculations (green dotted). Grey bands represent model uncertainty ($\pm 3\sigma$), while red bars (right axis) indicate absolute prediction errors. All error metrics are reported in weight percentage (wt.\%).}
    
    % MAPE ($\%$) of ML models on the test set for predicting \textbf{a} alite,\textbf{ b} belite, \textbf{c} ferrite with PP, KF, HM, and CO as input. The best-performing models are shown in bold. Comparing the accuracy of best performing ML against Bogue for predicting \textbf{d}, \textbf{e} alite \textbf{f},\textbf{g} belite and \textbf{h}, \textbf{i} ferrite. In subplots \textbf{d}, \textbf{f} and \textbf{h} the inset plot in the top left (for ML model) and bottom left (for Bogue) show the distribution of prediction errors ($\epsilon= \text{predicted - actual value}$) on the test data. The y-axis of the inset shows the frequency ($\nu$) of the prediction error. The red-shaded portion of the inset denotes the $ 95\%$ confidence interval ($\pm\: 2 \sigma$). The limits of the x-axis for insets represent the $99.9\%$ confidence interval ($\pm\: 4 \sigma$).  In subplots \textbf{e}, \textbf{g} and \textbf{i}, the grey region around the mean predictions is the model uncertainty ($\pm \: 3\sigma$). The red bars (calibrated to the right y-axis) represent the absolute error ( $|\text{predicted - actual}|$ ) of the model. }
    \label{Fig2new}
\end{figure}
Input feature selection critically influences model performance and practical utility in industrial settings. While comprehensive plant data theoretically offers maximum information content, a parsimonious model utilizing strategically selected input parameters may achieve comparable accuracy. The design of data-driven models were governed by two key aspects: accurate prediction of clinker mineralogy, and potential to implement model predictive. While there are limitations no the input parameters while considering the first aspect, the second requires the input to be directly controllable during the plant operation so as to control the clinker mineralogy and hence quality.

\subsubsection{Performance of ML models}
We categorized potential input combinations into two distinct classes: predictive control features (process parameters and raw materials) enabling real-time process optimization and post-production analysis features incorporating clinker oxide measurements. The latter, while potentially offering superior accuracy due to enriched chemical and process information, remains unsuitable for online process control due to inherent measurement delays and availability of the features post-production. To systematically evaluate these trade-offs, we constructed 15 distinct feature sets (Fig.~\ref{fig1new}a), encompassing various combinations of kiln feed characteristics, process parameters, hot meal properties, and clinker oxide compositions. This comprehensive approach enabled the identification of minimal yet sufficient feature sets for accurate phase prediction while maintaining practical applicability for process control.

We systematically evaluated eight machine learning architectures for clinker phase prediction using the complete feature set of 59 features from KF, HM, PP, and CO. Figure~\ref{Fig2new} compares the performance of linear models (linear regression, lasso, ridge, elastic net) against non-linear approaches (random forest, XGBoost, support vector regression (SVR), Gaussian process regression (GPR), neural network (NN)~\cite{karimipour_novel_2019,martinez-ramon_support_2006,ni_moving-window_2012,ni_recursive_2011,tealab_forecasting_2017}) across three primary clinker phases (see Methods for details). Table~\ref{Table_2} shows the performance of all the models for alite, belite and ferrite. Note that each model underwent rigorous cross-validation to ensure optimal hyperparameter selection and prevent overfitting.

For alite prediction (Fig.~\ref{Fig2new}a), linear models consistently demonstrated higher mean absolute percentage errors (MAPE), clustering near the radar plot periphery. This performance deficit persisted despite the inclusion of comprehensive process parameters, suggesting inherently non-linear relationships between input features and phase composition. This observation challenges the industry-standard Bogue equation, which assumes linear relationships between clinker oxides and phases. Non-parametric models—particularly NN, GPR, and SVR—achieved superior prediction accuracy across all three phases, with MAPEs of 1.24\%, 6.77\%, and 2.53\% for alite, belite, and ferrite, respectively (see Fig.~\ref{Fig2new}b,c).

\subsubsection{Benchmarking against Bogue equation}
To benchmark against industry standards, we compared our models against the plant-specific Bogue equation (Fig.~\ref{Fig2new}d-i), obtained directly from the domain experts in the specific plant. A two-month period, comprising January and February in 2020, was considered for this evaluation, which was kept unseen by the model and was excluded from the training data. Thus, this period is equivalent to an unseen operating period of the plant and serves as a true test of the ML model. The best-performing architectures corresponding to each of the minerals as identified from Fig.~\ref{Fig2new}a-c were used. ML model robustness was assessed through 20 independent training iterations with different random seeds. The best-performing architectures, namely, NN for alite, GPR for belite, SVR for ferrite, demonstrated remarkable improvements over Bogue predictions. Parity plots (Fig.\ref{Fig2new}d,f,h) reveal tighter clustering around the ideal prediction line, while temporal predictions over a two-month test period (Fig.~\ref{Fig2new}e,g,i) show that the models capture complex compositional dynamics with high fidelity. Error distributions of the models (inset histograms) confirm significantly reduced prediction variance compared to Bogue calculations. Further, statistical analysis validated the superior performance of ML models over traditional Bogue calculations. ML predictions demonstrated symmetric error distributions (Fig.~\ref{Fig2new}d), contrasting with Bogue's systematic biases. Neural network predictions for alite showed exceptional consistency across training and test sets, confirming robust generalization across the entire compositional range.

Notably, the ML models accurately tracked rapid compositional fluctuations while maintaining $\pm 3\sigma$ prediction confidence intervals (grey regions). Notably, some of the spikes in the alite, with compositional variations of $\sim$15 wt.\% in a day, were captured accurately by the ML model. In contrast, the Bogue equation exhibited systematic biases: overestimating alite content (MAPE: 7.79\% vs 1.24\%), underestimating belite composition (MAPE: 22.68\% vs 6.77\%), and severely misrepresenting ferrite concentrations (MAPE: 24.54\% vs 2.53\%). These results underscore the limitations of linear approximations in capturing complex clinker formation dynamics. More importantly, the results conclusively demonstrate the superior ability of ML models to predict the clinker compositions accurately despite the huge fluctuations, making them a promising tool for online model predictive control.

\begin{table}[h]
% \centering
\caption{\label{Table_2} Comparing the performance of ML models on the test set using MAE, $R^{2}$ and MAPE. The scores for the best, second best, and third best models are shown using bold, underlined, and italics, respectively }
\resizebox{\columnwidth}{!}{%
\begin{tabular}{|c|ccc|ccc|ccc|}
\hline
\multirow{2}{*}{\textbf{Models}} & \multicolumn{3}{c|}{\textbf{Alite}} & \multicolumn{3}{c|}{\textbf{Belite}} & \multicolumn{3}{c|}{\textbf{Ferrite}} \\ \cline{2-10} 
 & \textbf{MAE(wt. \%)} & \textbf{R$^{2}$} & \textbf{MAPE (\%)} & \textbf{MAE(wt. \%)} & \textbf{R$^{2}$} & \textbf{MAPE (\%)} & \textbf{MAE(wt. \%)} & \textbf{R$^{2}$} & \textbf{MAPE (\%)} \\ \hline
\textbf{Linear regression} & 1.11 & 0.79 & 1.86 & 1.23 & 0.76 & 8.70 & 0.40 & 0.63 & 2.85 \\
\textbf{Lasso} & 1.11 & 0.79 & 1.86 & 1.23 & 0.76 & 8.70 & 0.40 & 0.63 & 2.85 \\
\textbf{Ridge} & 1.13 & 0.78 & 1.89 & 1.24 & 0.75 & 8.78 & 0.41 & 0.63 & 2.86 \\
\textbf{Elastic net} & 1.12 & 0.78 & 1.87 & 1.23 & 0.76 & 8.73 & 0.40 & 0.63 & 2.85 \\
\textbf{Random forest} & 1.10 & 0.79 & 1.82 & 1.15 & 0.78 & 8.24 & 0.39 & 0.64 & 2.77 \\
\textbf{XGBoost} & 1.01 & 0.82 & 1.69 & 1.03 & 0.83 & 7.34 & 0.39 & 0.65 & 2.76 \\
\textbf{SVR} & \textit{0.96} & \textit{0.84} & \textit{1.62} & \textit{0.99} & \textit{0.84} & \textit{7.12} & \textbf{0.36} & \textbf{0.69} & \textbf{2.53} \\
\textbf{GPR} & {\underline {0.96}} & {\underline{0.84} } & {\underline{1.61} } & \textbf{0.96} & \textbf{0.86} & \textbf{6.77} & {\underline{0.36} } & {\underline{} 0.70} & {\underline{2.54} } \\
\textbf{NN} & \textbf{0.75} & \textbf{0.90} & \textbf{1.24} & {\underline{0.98} } & {\underline{0.84} } & {\underline{6.96} } & \textit{0.40} & \textit{0.65} & \textit{2.80} \\ \hline
\end{tabular}%
}
\end{table} 

\begin{figure}[h!tbp]
    \centering
    \resizebox{0.7\textwidth}{!}{%
    \includegraphics{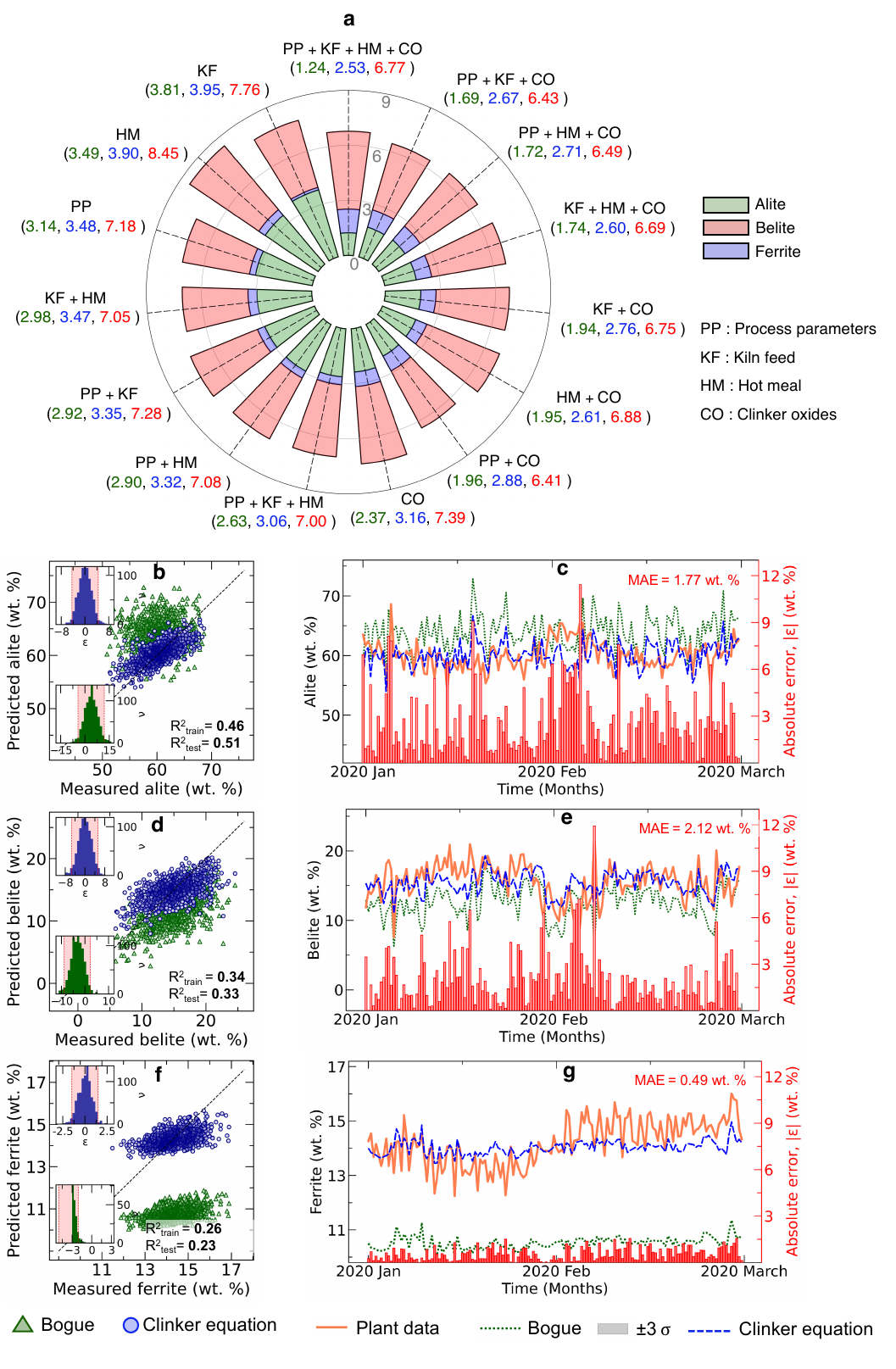}}
    \caption{\textbf{Comparison with Bogue equation.} a, MAPE of optimal machine learning models (Neural Network for alite, Gaussian Process Regression for belite, Support Vector Regression for ferrite) across 15 combinations of input features: process parameters (PP), kiln feed (KF), hot meal (HM), and clinker oxides (CO). Values in parentheses represent MAPE (\%) for alite (green), ferrite (blue), and belite (red) predictions. b-g, Performance evaluation of plant-specific clinker equations against standard Bogue calculations. Parity and temporal plots comparing predicted versus measured compositions for b,c, alite, d,e, belite, and f,g, ferrite, respectively. Inset histograms show error distributions for clinker equations (top) and Bogue calculations (bottom). The temporal evolution of predictions is over a two-month test period showing plant data (red), clinker equation predictions (blue dashed), and Bogue calculations (green dotted). Grey bands represent model uncertainty (±3$\sigma$), while red bars (right axis) indicate absolute prediction errors. Training ($R^2_{train}$) and test ($R^2_{test}$) set performance metrics demonstrate superior accuracy of plant-specific equations over traditional Bogue calculations. All compositions and errors are reported in weight percentage (wt.\%).}
    \label{Fig 7}
\end{figure}

\textbf{Input feature pruning.}
To identify the best combination of input features that balance between clinker prediction accuracy and predictive control, we systematically evaluated model performance across 15 distinct feature combinations through a radial visualization (Fig.~\ref{Fig 7}a). The full-feature models (PP+KF+HM+CO) achieved optimal accuracy with MAPEs of 1.24\%, 2.53\%, and 6.77\% for alite, ferrite, and belite, respectively. However, these post-production predictions, while accurate, offer limited utility for real-time process control. Specifically, any model with CO as input features inhibit predictive control as CO measurements are obtained only post-production. Thus, the remaining set of 7 models, with input features excluding CO were analyzed. 

We observed that models without CO exhibit reasonable performance, albeit slightly poorer than those with CO as input feature. However, all the models performed better than the Bogue equation. Notably, even reduced-feature ML models using only process parameters outperformed Bogue calculations: alite (MAPE: 3.14\% vs 7.79\%), ferrite (3.48\% vs 24.54\%), and belite (7.18\% vs 22.68\%). This suggests that a predictive control implemented purely based on PP can outperform those based Bogue, not to mention that Bogue requires the clinker oxide compositions which can be obtained only post-production. The gradual degradation in prediction accuracy with feature reduction is clearly visualized in the radial plot, providing process engineers with a quantitative framework for feature selection based on specific accuracy requirements.

\textbf{Plant specific equations.} To establish a fair comparison between Bogue equation, we developed plant-specific linear regression models (clinker equations) using identical input parameters as Bogue equations (Fig.~\ref{Fig 7}b-g) (see Appendix~\ref{clinker equations}). These tailored equations demonstrated marked improvements over standard Bogue calculations, particularly for alite ($R^2_{test}=0.51$ vs $R^2_{test}=0.23$) and belite ($R^2_{test}=0.33$ vs $R^2_{test}=0.26$) predictions. The temporal evolution plots (Fig.\ref{Fig 7}c,e,g) reveal significantly reduced mean absolute errors: 1.77 wt.\%, 2.12 wt.\%, and 0.49 wt.\% for alite, belite, and ferrite respectively. This analysis conclusively demonstrates that even simplified plant-specific models offer substantially more reliable quality control metrics than traditional Bogue calculations. Thus, instead of relying on traditional Bogue equations, having simplified plant-specific equations obtained purely in a data-driven fashion can serve as a better performance indictor to be used for quality control. Detailed equations and their derivations are presented in~\ref{clinker equations}.

% \newgeometry{bottom=1cm}
\begin{figure}[h!tbp]
\resizebox{\textwidth}{!}{%
    \centering
    \includegraphics[scale=1.0]{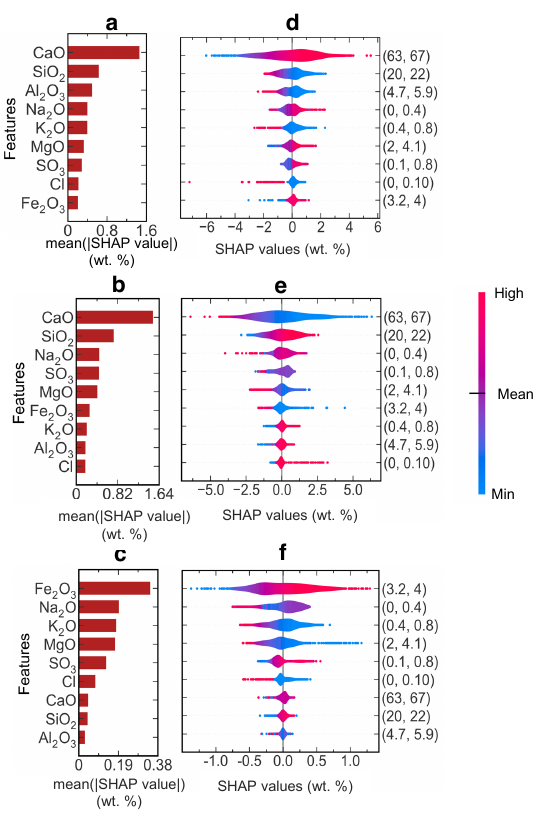}}
    \caption{\textbf{Feature attribution analysis of clinker phase predictions using SHAP.} a-c, Hierarchical ranking of clinker oxide contributions to phase predictions for alite, belite, and ferrite, respectively. Bar lengths indicate mean absolute SHAP values (wt.\%), representing averaged feature impact across the test dataset. d-f, Corresponding beeswarm plots revealing the directional influence of each oxide on phase formation. SHAP values (x-axis) indicate deviation from mean phase composition, with positive values suggesting increased formation. Color gradient (blue to red) represents oxide concentration from minimum to maximum, with point density indicating frequency of occurrence. Numbers in parentheses show oxide composition ranges (wt.\%). CaO demonstrates a dominant positive correlation with alite formation, while SiO$_2$ shows a strong negative influence, aligning with established clinker chemistry.}
    % SHAP Bar plots for \textbf{a} alite, \textbf{b} belite, and \textbf{c} ferrite show the average impact of clinker oxides on model predictions. The clinker oxides are arranged from top to bottom in order of decreasing impact on the model predictions. $Mean \:|\: SHAP\:values\:|$ (represented on the X-axis) for oxide is computed by averaging the absolute SHAP values for that oxide across the entire test set. SHAP Beeswarm plots for \textbf{d} alite, \textbf{e} belite, and \textbf{f} ferrite show the directionality of an oxide’s impact on the model prediction. An increase in SHAP values (shown on the X-axis) indicates an increase in the clinker phase and vice versa. The color and thickness of the curve corresponding to each oxide represent the value and density of that oxide in the test set. The high value of an oxide is represented by red, and the low by blue. The range of the clinker oxides is shown on the right Y-axis as $(min, max)$ in $wt.\:\%.$ Note that $0$ on the x-axis represents the mean of the clinker phases in the test set, and the increase (or decrease) of phases described in the SHAP is with respect to the mean of respective phases in the test set.}
    \label{fig:last figure}
\end{figure}
% \restoregeometry
\subsection{Interpreting the ML models}
To bridge the gap between model performance and domain expertise, we employed SHAP (SHapley Additive exPlanations) analysis to quantify feature contributions and their directional impact on clinker phase predictions. We focused our analysis on clinker oxide (CO) features, excluding process parameters (PP) due to their complex interdependencies, to enable direct comparison with established mineralogical understanding.

The hierarchical influence of clinker oxides emerges distinctly in Fig.~\ref{fig:last figure}a-c. For alite prediction, CaO and SiO$_{2}$ demonstrate dominant influence, with mean absolute SHAP values of 1.6 and 0.8 wt.\%, respectively. The beeswarm visualization (Fig.~\ref{fig:last figure}d) reveals that increased CaO content (63-67 wt.\%) positively correlates with alite formation, while elevated SiO$_{2}$ levels (20-22 wt.\%) exhibit negative correlation, aligning with classical clinker chemistry principles.

Belite predictions (Fig.~\ref{fig:last figure}b,e) show similar oxide dependencies, though with distinct quantitative relationships. CaO maintains primary influence (mean$\:|\:\textrm{SHAP}\:\textrm{values}\:|$ = 1.64 wt.\%), followed by SiO${_2}$ and Na${_2}$O. The ferrite phase demonstrates unique sensitivity to Fe${_2}$O${_3}$ content (mean$\:|\:\textrm{SHAP}\:\textrm{values}\:|$ = 0.38 wt.\%), with alkali oxides (Na${_2}$O, K${_2}$O) exhibiting secondary influence (Fig.~\ref{fig:last figure}c,f), as Fe${_2}$O${_3}$ primarily governs and represents the ferrite phase formation.

It is also worth noting that both the importance of the features and directionality of their influence (positive vs negative) is congruent with the plant-specific and the Bogue equations. Thus, the SHAP-derived relationships corroborate that the relationship learned by the ML models are congruent with the established domain knowledge while providing quantitative insights into feature interactions. The analysis confirms that our ML models capture fundamental physicochemical relationships governing clinker phase formation, enhancing their credibility for industrial deployment.

\section{Conclusions}
The present work establishes a new paradigm in cement manufacturing through precise mineralogical predictions from large-scale plant data for the first time, to the best of our knowledge. Our computational framework significantly outperforms conventional Bogue equations across all metrics while maintaining exceptional performance even with minimal input parameters, facilitating both instantaneous monitoring and retrospective analysis. Moreover, the present work can provide on-the-fly predictions of the clinker compositions, while Bogue equation is a post-mortem analysis based on the clinker oxides.

The results presented provide key insights for industrial cement production in three critical dimensions. First, ML models can provide accurate estimates for clinker phase prediction, a feat that has not been possible thus far based on large-scale operational plant data with huge fluctuations. Second, ML-based digital twins utilizing solely operational parameters and raw material compositions can provide pre-production estimations, enabling the development of robust quality assurance protocols and potential for online process control. Third, facility-specific equations derived from operational measurements markedly exceed standard calculations, offering pragmatic intermediate solutions during digital transformation initiatives.

Beyond clinker prediction, the present work illuminates several frontiers in sustainable manufacturing. Integration with supervisory systems could enable autonomous optimization of product specifications while reducing thermal energy requirements. The framework naturally extends toward predicting additional material characteristics, such as 28-day strength, and accommodating alternative constituents, such as alternative fuels and raw materials in cement production. This is crucial in cement manufacturing as the sector alone contributes toward $\sim$10\% of the global carbon emissions. Thus, moving towards data-driven digital twins in cement manufacturing holds the potential to improve existing industrial systems with little changes, potentially accelerating progress toward carbon-neutral manufacturing practices.

\section*{Data availability}
The raw data used in this study was obtained from an industry partner under a confidentiality agreement with the Global Cement and Concrete Research Network (GCCRN). The identity of the industry partner and the raw dataset cannot be disclosed due to proprietary restrictions. Consequently, the data is not publicly available. Any inquiries regarding the data can be directed to the corresponding author, subject to approval from the industry partner and GCCRN.

\section*{Code availability}
All the source codes developed as part of this work are publicly available at the following \href{https://github.com/M3RG-IITD/Industry_Clinker_Mineralogy.git}{Git repository}.

\section*{Acknowledgements}
The authors acknowledge the funding and data collection support by Innovandi - The Global Cement and Concrete Research Network (GCCRN). The authors also acknowledge the high-performance computing (HPC) platform at the Indian Institute of Technology Delhi for the computational and storage resources.

\section*{CRediT authorship contribution statement}
\begin{figure}[H] 
    \begin{flushleft}
        \resizebox{0.5\textwidth}{!}{%
        \includegraphics[scale=0.6]{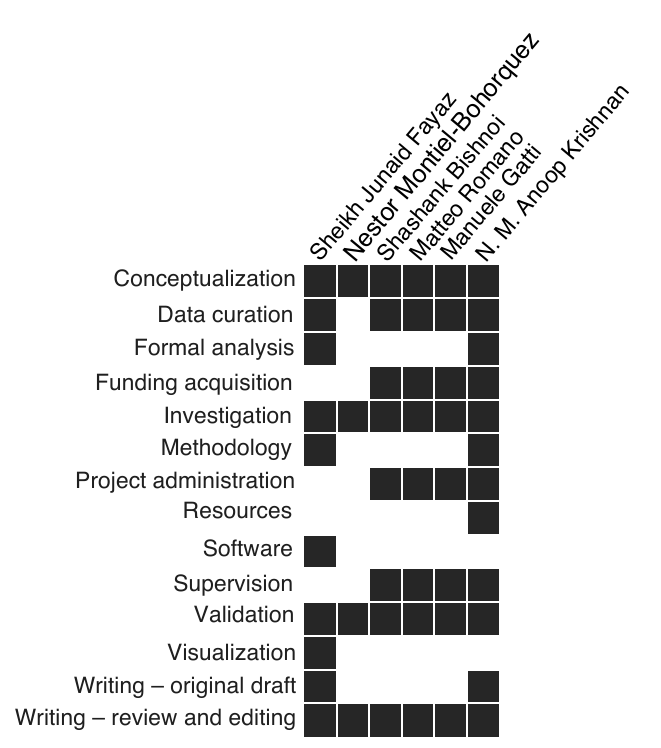}}
    \end{flushleft}
\end{figure}

\section*{Competing interests}
The authors declare no competing interests.

\bibliographystyle{elsarticle-num}
\bibliography{Innovandi_Clk_pred}

\newpage
\appendix
\vspace{2pt}
\begin{center}
\parbox{0.99\textwidth}{\centering \Large Industrial-scale Prediction of Cement Clinker Phases using Machine Learning}
\end{center}
\vspace{2pt}
\begin{center}
\parbox{1.0\textwidth}{\centering \normalsize Sheikh Junaid Fayaz$^{a}$, Néstor Montiel-Bohórquez$^{b}$, Shashank Bishnoi$^{a}$, Matteo Romano$^{b}$, Manuele Gatti$^{b}$, N. M. Anoop Krishnan$^{a}$}
\end{center}
\vspace{2pt}
\begin{center}
    \section*{Supplementary Materials}
\end{center}

\section{Data}\label{appendix a data}
The list of 34 process parameters that were considered for this study are reported below in Table\ref{Table_A1}. The description of composition data reported in DB2 is also shown in Table~\ref{Table_A2}. It is important to mention that the XRD measurements for  aluminate, freelime and other minor phases were not available in the data collected from the plant. The alite and belite compositions reported in Table~\ref{Table_A1} is the sum of constituent polymorphs. Also, the raw data collected from the plant has some negative values as well for some of the composition entries. For instance, see minimum value for kiln feed Cl in Table~\ref{Table_A2}. In total, there were 44 rows consisting of such inconsistent data entries. These were removed during the datapreprocessing as mentioned in Table~\ref{Table_A3}.

\begin{table}[h]
\caption{\label{Table_A1} Description of process parameters reported in DB1.}
\centering
\resizebox{\columnwidth}{!}{%
\begin{tabular}{ccccc}
\hline
\textbf{Notation} & \textbf{Parameter} & \textbf{Description} & \textbf{Unit} & \textbf{Range} \\
\hline\\
$P_{1}$  & Exit temp C1A                          & Gas outlet temperature (stage 1A cyclone)     & °C                        & 0 - 600        \\
$P_{2}$  & Exit temp C1B                          & Gas outlet temperature (stage 1B cyclone)     & °C                        & 0 - 600        \\
$P_{3}$  & Exit temp C2                           & Gas outlet temperature (stage 2 cyclone )     & °C                        & 0 - 800        \\
$P_{4}$  & Exit temp C3                           & Gas outlet temperature (stage 3 cyclone )     & °C                        & 0 - 900        \\
$P_{5}$  & Exit temp C4                           & Gas outlet temperature (stage 4 cyclone )     & °C                        & 0 - 1000       \\
$P_{6}$  & Exit temp C5                           & Gas outlet temperature (stage 5 cyclone )     & °C                        & 0 - 1200       \\
$P_{7}$  & PH gas outlet temp                     & Preheater gas outlet temperature                & °C                        & 0 - 600        \\
$P_{8}$  & O$_{2}$ in raw gas at PH outlet             & O$_{2}$ in raw gas measured at preheater outlet & \%                        & 0 - 25         \\
$P_{9}$  & Temp of KF to PH                       & Raw mill outlet temperature                         & °C                        & 0 - 150        \\
$P_{10}$ & KF flow rate to PH                     & Total kiln feed includes fly ash and dust       & tph                       & 0 - 550.0      \\
$P_{11}$ & Flue gas exit temperature              & Calciner exit duct centered south temp         & °C                        & 0 - 1370       \\
$P_{12}$ & Flue gas O$_{2}$ content at calciner outlet & Dry/wet basis not reported                     & \%                        & 0 - 25         \\
$P_{13}$ & Solid's outlet temp at kiln inlet     & Measured at rotary kiln inlet                  & °C                        & 700 - 1500     \\
$P_{14}$ & Calciner's fuel consumption            & Calciner coal feeding                          & tph                       & 0 - 36         \\
$P_{15}$ & O$_{2}$ in the kiln gas at kiln inlet       & Measured at rotary kiln inlet                  & \%                        & 0 - 25         \\
$P_{16}$ & HM temp of lowest cyclone              & Same as Stage 5 outlet temperature             & °C                        & 0 - 1200       \\
$P_{17}$ & Total cooling air                      & Total cooling air                              & Am$^{3}$/h                & 0 - 630000     \\
$P_{18}$ & III Air temp at CC outlet              & Heat exchanger inlet temperature \#1           & °C                        & 0 - 650        \\
$P_{19}$ & Clinker production                     & Clinker flow rate (calculated)                 & tph                       & 0 - 100        \\
$P_{20}$ & Total flow of gas entering GCT         & GCT wet air flow                               & CFM                       & 0 - 3000       \\
$P_{21}$ & Temp of gas entering GCT               & GCT inlet temperature                          & °C                        & 0 - 500        \\
$P_{22}$ & Spray water used in GCT                & GCT water flow                                 & m$^{3}$/h                 & 0 - 70         \\
$P_{23}$ & GCT outlet temp                        & GCT outlet temperature                         & °C                        & 0 - 500        \\
$P_{24}$ & Temp of gas entering the main fan      & Measured at main fan inlet                     & °C                        & 0 - 600        \\
$P_{25}$ & Raw mill electric consumption               & Raw mill motor power                           & kW                        & 0 - 6000       \\
$P_{26}$ & Pre-calciner outlet pressure           & Preheater stage 5 pressure above kiln inlet    & mbar                      & -50 - 5            \\
$P_{27}$ & Exit pressure C1A                      & Gas outlet pressure (stage 1A cyclone)       & mbar                      & -120 -10           \\
$P_{28}$ & Exit pressure C1B                      & Gas outlet pressure (stage 1B cyclone)       & mbar                      & -120 - 10           \\
$P_{29}$ & Exit pressure C2                       & Gas outlet pressure (stage 2 cyclone )       & mbar                      & -70 - 10            \\
$P_{30}$ & Exit pressure C3                       & Gas outlet pressure (stage 3 cyclone )       & mbar                      & -60 - 10           \\
$P_{31}$ & Exit pressure C4                       & Gas outlet pressure (stage 4 cyclone )       & mbar                      & -50 - 5            \\
$P_{32}$ & Exit pressure C5                       & Gas outlet pressure (stage 5 cyclone )       & mbar                      & -40 - 5            \\
$P_{33}$ & Raw mill fan inlet pressure at pt 1    & Raw mill inlet pressure 1                      & mbar                      & -15 - 5           \\
$P_{34}$ & Raw mill fan inlet pressure at pt 2    & Raw mill inlet pressure 2                      & mbar                      & -15 - 5            \\ \\
\hline
\end{tabular}%
}
\end{table}

\begin{figure}[h!tbp]
\resizebox{\textwidth}{!}{%
    \centering
    \includegraphics[scale=1.0]{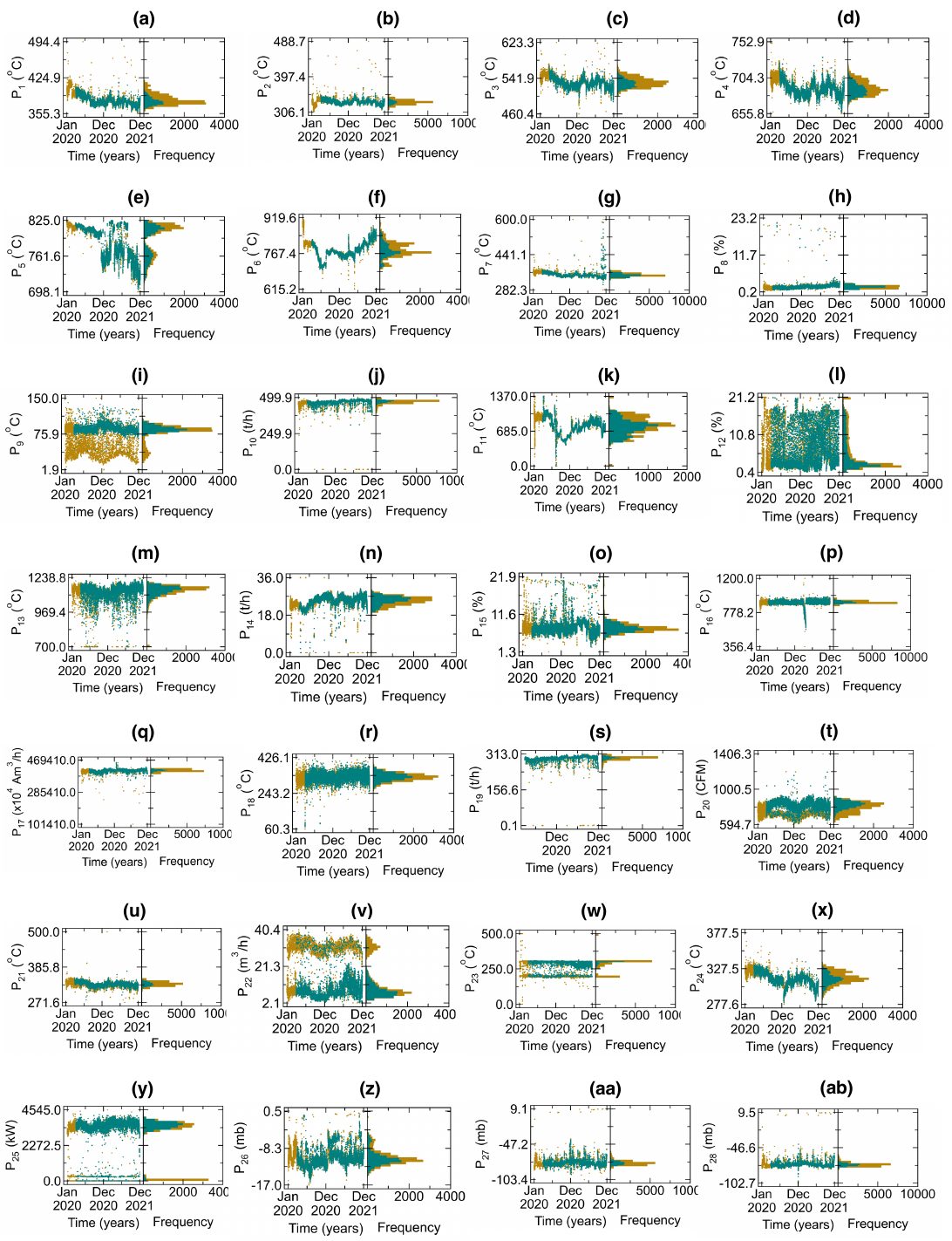}}
    \caption{\textbf{Timeseries and distribution analysis of all process parameters used in the study}. Raw plant data is represented in yellow, while pre-processed data is shown in green. The notation for each process parameter on the y-axis corresponds to the labels provided in Table \ref{Table_A3} }
    \label{dist_PP_1}
\end{figure}

\begin{table}[h]
\caption{\label{Table_A2} Details of the KF, HM and clinker composition data reported in DB2}
\centering
\resizebox{\columnwidth}{!}{%
\begin{tabular}{ccccccccc}
\hline
\textbf{Database}                   & \textbf{Composition} & \textbf{Unit} & \textbf{Min} & \textbf{Max} & \textbf{Mean} & \textbf{\begin{tabular}[c]{@{}c@{}}Standard \\ deviation\end{tabular}} & \textbf{\begin{tabular}[c]{@{}c@{}}Measurement \\ technique\end{tabular}} & \textbf{\begin{tabular}[c]{@{}c@{}}Measurement \\ point\end{tabular}} \\
\hline
\multirow{9}{*}{\textbf{Kiln feed}} & CaO                  & wt. \%        & 38.17            & 44.30            & 42.43         & 0.36                       & XRF                            & \multirow{9}{*}{\text{\begin{tabular}[c]{@{}c@{}}Before the  \\ preheater tower\end{tabular}}} \\
                                    & SiO$_{2}$            & wt. \%        & 10.36            & 15.04            & 13.02         & 0.26                       & XRF                            &                                      \\
                                    & Al$_{2}$O$_{3}$      & wt. \%        & 2.38             & 4.01             & 2.96          & 0.11                       & XRF                            &                                      \\
                                    & Fe$_{2}$O$_{3}$      & wt. \%        & 1.86             & 2.71             & 2.11          & 0.05                       & XRF                            &                                      \\
                                    & MgO                  & wt. \%        & 1.31             & 2.71             & 2.08          & 0.19                       & XRF                            &                                      \\
                                    & SO$_{3}$             & wt. \%        & 0.03             & 0.61             & 0.13          & 0.06                       & XRF                            &                                      \\
                                    & K$_{2}$O             & wt. \%        & 0.31             & 0.55             & 0.43          & 0.03                       & XRF                            &                                      \\
                                    & Na$_{2}$O            & wt. \%        & 0.00             & 0.18             & 0.06          & 0.02                       & XRF                            &                                      \\
                                    & Cl                   & wt. \%        & -0.01            & 0.10             & 0.02          & 0.01                       & XRF                            &                                      \\
\hline
\multirow{7}{*}{\centering\textbf{Hot meal}}  & SO$_{3}$       & wt. \%        & -0.02            & 2.97             & 0.88          & 0.18                       & XRF                            & \multirow{7}{*}{\centering\text{\begin{tabular}[c]{@{}c@{}}Before entering \\ the rotary kiln \\ (calciner exit) \end{tabular}}}  \\
                                    & K$_{2}$O             & wt. \%        & 0.38             & 4.00             & 1.51          & 0.22                       & XRF                            &                                      \\
                                    & Na$_{2}$O            & wt. \%        & 0.07             & 11.73            & 0.17          & 0.11                       & XRF                            &                                      \\
                                    & Cl                   & wt. \%        & 0.00             & 3.25             & 0.64          & 0.20                       & XRF                            &                                      \\
                                    & Alite                & wt. \%        & 0.00             & 62.75            & 2.22          & 1.48                       & XRD                            &                                      \\
                                    & Belite               & wt. \%        & 0.00             & 23.76            & 2.69          & 1.21                       & XRD                            &                                      \\
                                    & Ferrite              & wt. \%        & 0.00             & 20.45            & 1.31          & 0.77                       & XRD                            &                                      \\
\hline
\multirow{12}{*}{\textbf{Clinker}}  & CaO                  & wt. \%        & 57.71            & 66.75            & 64.62         & 0.42                       & XRF                            & \multirow{12}{*}{\text{\begin{tabular}[c]{@{}c@{}} Exit of \\ clinker cooler \\\ outlet\end{tabular}}}  \\
                                    & SiO$_{2}$            & wt. \%        & 18.66            & 27.10            & 20.89         & 0.29                       & XRF                            &                                      \\
                                    & Al$_{2}$O$_{3}$      & wt. \%        & 4.57             & 7.33             & 5.24          & 0.17                       & XRF                            &                                      \\
                                    & Fe$_{2}$O$_{3}$      & wt. \%        & 3.06             & 4.42             & 3.55          & 0.11                       & XRF                            &                                      \\
                                    & MgO                  & wt. \%        & 1.90             & 4.17             & 3.25          & 0.36                       & XRF                            &                                      \\
                                    & SO$_{3}$             & wt. \%        & 0.07             & 1.43             & 0.50          & 0.10                       & XRF                            &                                      \\
                                    & K$_{2}$O             & wt. \%        & 0.41             & 0.87             & 0.60          & 0.06                       & XRF                            &                                      \\
                                    & Na$_{2}$O            & wt. \%        & 0.00             & 0.39             & 0.08          & 0.04                       & XRF                            &                                      \\
                                    & Cl                   & wt. \%        & -0.01            & 0.32             & 0.01          & 0.01                       & XRF                            &                                      \\
                                    & Alite                & wt. \%        & 0.00             & 78.33            & 60.03         & 3.68                       & XRD                            &                                      \\
                                    & Belite               & wt. \%        & 0.00             & 40.61            & 15.11         & 3.43                       & XRD                            &                                      \\
                                    & Ferrite              & wt. \%        & 0.00             & 42.23            & 14.31         & 1.03                       & XRD                            &                                      \\
\hline
\end{tabular}%
}
\end{table}
The raw dataset reported a total of 14,985 clinker compositions. Two hundred-seven clinker compositions did not have the corresponding PP, KF, and/or HM measurements. Consequently, these clinker compositions were eliminated. Additionally, there were 3,170 missing entries in the raw data. The raw dataset also indicates some negative compositions. (Refer to Table \ref{Table_A2} for the minimum values of kiln feed Cl, hot meal SO$_{3}$ and clinker Cl). A total of 5,857 negative entries were removed from DB1.  The data loss throughout the various stages of data preparation is detailed in Table \ref{Table_A3}. A comprehensive representation of the preprocessed data in comparison to the raw data is included in Fig\ref{dist_PP_1} to Fig \ref{hrm dist}.

\begin{figure}[h!tbp]
\resizebox{\textwidth}{!}{%
    \centering
    \includegraphics[scale=1.0]{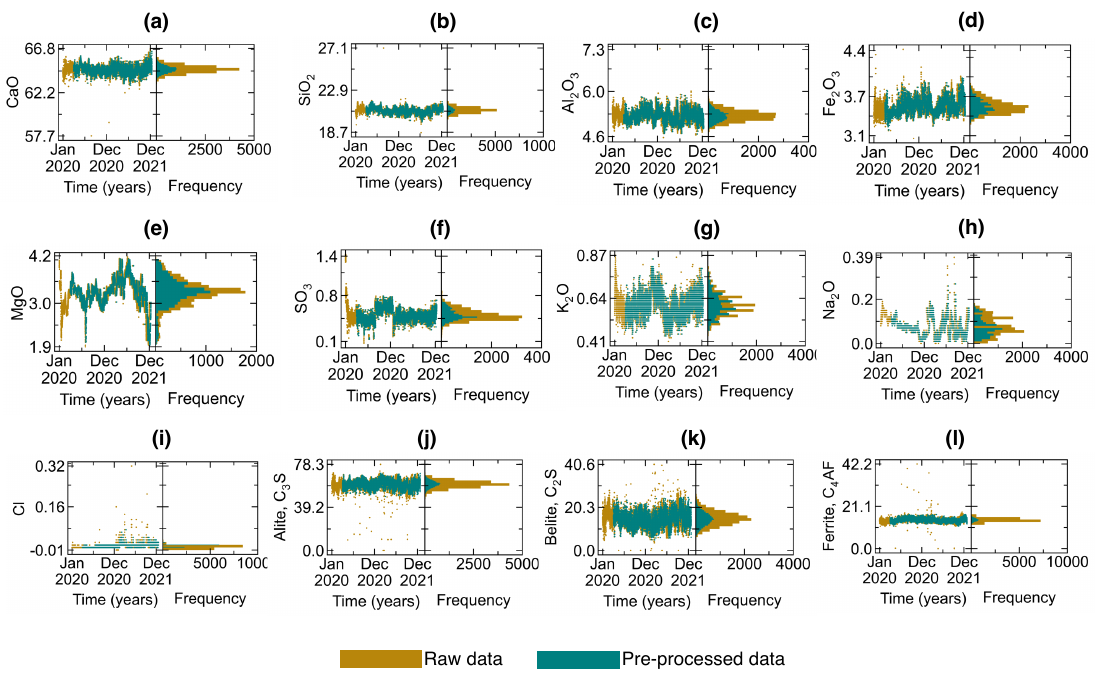}}
    \caption{\textbf{Timeseries and distribution analysis of all compositions reported in the clinker database from the plant}. Raw plant data is represented in yellow, while pre-processed data is shown in green. The left and right subplots show scatter plots and histograms, respectively, for each of the reported clinker compositions.
    }
    \label{clinker dist}
\end{figure}

\begin{figure}[h!tbp]
\resizebox{\textwidth}{!}{%
    \centering
    \includegraphics[scale=1.0]{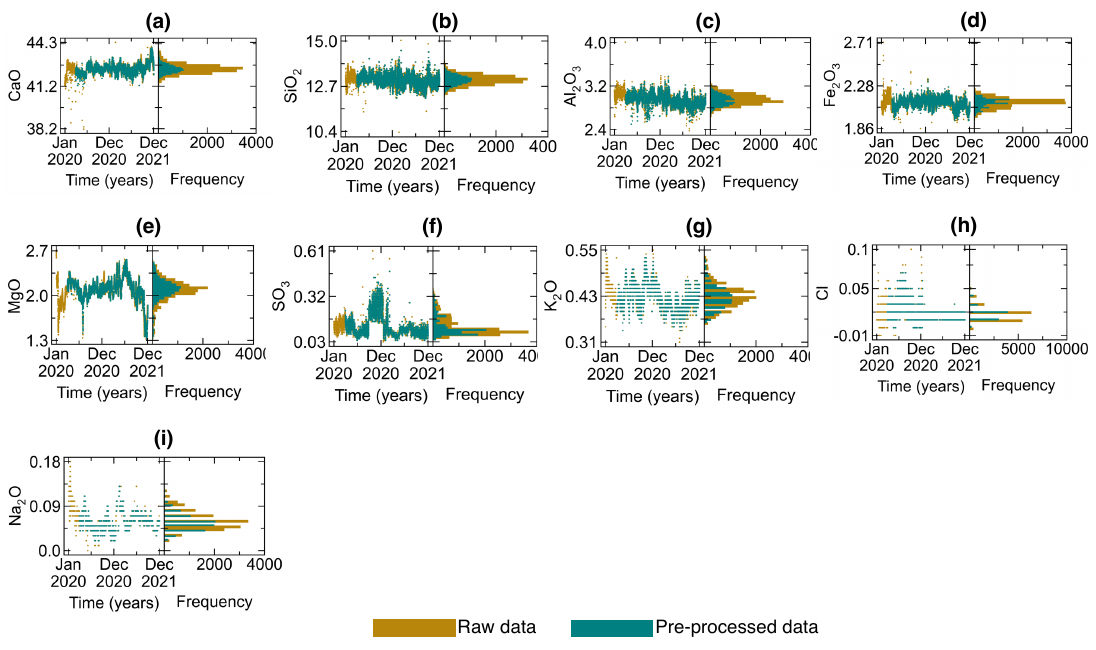}}
    \caption{\textbf{Timeseries and distribution analysis of all compositions reported in the feed raw meal database from the plant}. Raw plant data is represented in yellow, while pre-processed data is shown in green. The left and right subplots show scatter plots and histograms, respectively, for each of the reported kiln feed compositions.
    }
    \label{frm dist}
\end{figure}

\begin{figure}[h!tbp]
\resizebox{\textwidth}{!}{%
    \centering
    \includegraphics[scale=1.0]{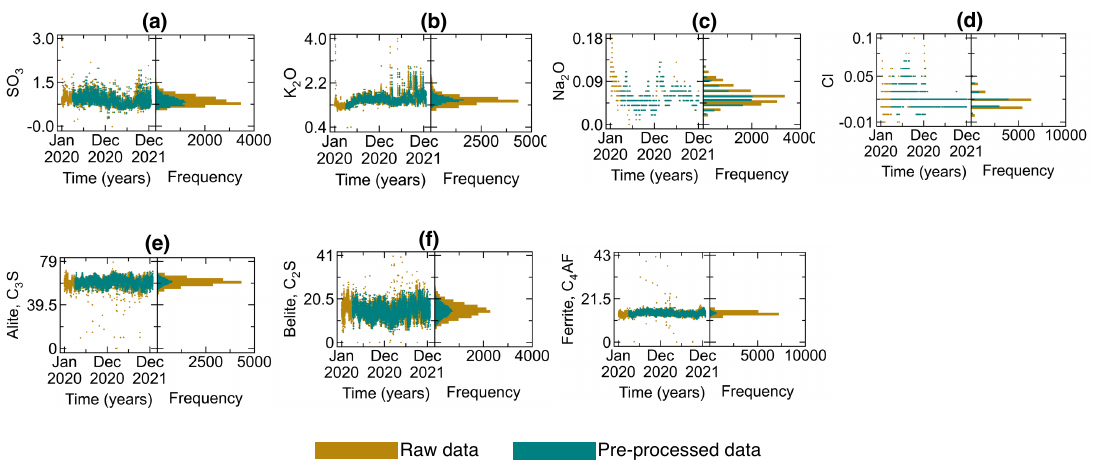}}
    \caption{\textbf{Timeseries and distribution analysis of all compositions reported in the hot meal database from the plant}. Raw plant data is represented in yellow, while pre-processed data is shown in green. The left and right subplots show scatter plots and histograms, respectively, for each of the reported hot meal compositions.
    }
    \label{hrm dist}
\end{figure}

\begin{table}[h]
\caption{\label{Table_A3} \centering{Data loss through subsequent stages of preprocessing. The percentage  calculation for  the data rows removed is relative to the size of the raw clinker composition database, i.e., 14,985}}
\centering
\resizebox{\columnwidth}{!}{%
\begin{tabular}{ccccccccc}
\hline
\multirow{2}{*}{\textbf{Operation}}                                             & \multirow{2}{*}{\textbf{\begin{tabular}[c]{@{}c@{}} Data rows \\ removed \end{tabular}}}                & \multicolumn{2}{c}{\textbf{Total data size}}         \\
                                                                                &                                                            & \textbf{Before} & \textbf{After} \\
\hline \\
{\text{\begin{tabular}[c]{@{}c@{}} Matching clinker compositions with \\ corresponding KF, HM and PP measurements \end{tabular}}}     & \begin{tabular}[c]{@{}c@{}}207 \\ (1.4 \%)\end{tabular}    & 14,985                    & 14,778                   \\ \\
Removing data rows with missing entries                                         & \begin{tabular}[c]{@{}c@{}}3,170 \\ (21.2 \%)\end{tabular} & 14,779                    & 11,608                   \\ \\
Removing negative compositions from DB1                                          & \begin{tabular}[c]{@{}c@{}}44\\ (0.3 \%)\end{tabular}      & 11,608                    & 11,564                   \\ \\
{\text{\begin{tabular}[c]{@{}c@{}} Outlier removal: eliminating data outside\\the range of 0.01 - 99.99 percentile \end{tabular}}}   & \begin{tabular}[c]{@{}c@{}}2,910\\ (19. 4\%)\end{tabular}  & 11,564                    & 8,654     
\\
\hline
\end{tabular}%
}
\end{table}

\section{ML algorithms} \label{ML models}
The following ML algorithms were implemented for modeling the clinker mineralogy in this study:
\subsection{Linear Regression (LR)}
LR assumes a linear relationship between the predictors and the output. Thus making it unsuitable for application to nonlinear systems such as the operation of a cement kiln. For a  data, ${D_{t}}$
\begin{equation}
    {D_{t}} = \left \{
    \underbrace{\begin{bmatrix}
        x_{1}^{1} & x_{2}^{1} & \ldots & x_{n}^{1} \\
        x_{1}^{2} & x_{2}^{2} & \ldots & x_{n}^{2} \\
        \vdots & \vdots & \ddots & \vdots \\
        x_{1}^{p} & x_{2}^{p} & \ldots & x_{n}^{p} 
    \end{bmatrix}}_\textbf{X}
    , 
    \underbrace{\begin{bmatrix}
        y^{1}\\
        y^{2} \\
        \vdots \\
        y^{p}
    \end{bmatrix}}_\textbf{Y}
    \right \}
    \label{training data}
\end{equation}
 the target variable can be represented using linear regression in the following form \cite{james_introduction_2021}:
\begin{equation}
   Y=\theta_0 + \sum_{j=1}^{n} \theta_j X_j + \epsilon 
\end{equation} 
 where $X_j= [x_j^1,x_j^2,\ldots, x_j^i,\ldots, x_j^p]^{T}$ , $\theta_0$ is the intercept, $\theta_j$ are the weights associated with each of the variable $X$ and $\epsilon$ is the uncertainty error. $\theta= [\theta_1, \theta_2, \ldots, \theta_j,\ldots, \theta_n ]^{T}$  represents the model parameters which are determined by fitting a hyperplane to the data ${D_{t}}$ by minimizing the loss function which in this case is the residual sum of squares (RSS)\cite{bishop_pattern_2006}:
 \begin{equation}
   \textnormal{RSS}(\theta)= \sum_{i=1}^{p}[y^i-(\theta_0+ \sum_{j=1}^{n} \theta_j X_j)]^2 
\end{equation} 
The predictor variables having more influence on the target output have higher weights and vice versa. However, in the linear regression method, weights for all input variables are non-zero, no matter how insignificant the contribution of the variable might be. 

\subsection{Lasso regression}
Overfitting \cite{ghojogh_theory_2023} is a frequently encountered problem in ML that arises due to the large variance of the ML models when tested on unseen data. Such variance can be reduced by applying appropriate penalties in the loss function. Lasso, ridge, and elastic net regression algorithms are tweaked variants of linear regression algorithms wherein different penalizing measures are used in the loss functions to reduce model variance. In lasso regression, the weights of the input variables are penalized with a factor called the L1 norm, which shrinks the weight of insignificant variables to zero. The L1 norm is applied to the loss function (RSS) as follows \cite{anoop_krishnan_predicting_2018}:
\begin{equation}
   \hat{\theta}_{lasso}=  \operatorname*{argmin}_\theta \left\{\sum_{i=1}^{p}[y^i-(\theta_0+ \sum_{j=1}^{n} \theta_j X_j)^2]+\lambda_{1} \sum_{j=1}^{n} \lvert \theta_j \rvert \right\}
\end{equation} 
where $\lambda_{1}$ is a hyperparameter.
\subsubsection{Ridge regression} 
Ridge regression, like lasso, is a regularized version of LR. However, it uses the $L2$ norm on the squared weights as follows:
\begin{equation}
   \hat{\theta}_{ridge}=  \operatorname*{argmin}_\theta \left\{\sum_{i=1}^{p}[y^i-(\theta_0+ \sum_{j=1}^{n} \theta_j X_j)^2]+\lambda_{2} \sum_{j=1}^{n} \lvert \theta_j \rvert^{2} \right\}
\end{equation} 
where $\lambda_{2}$ is a hyperparameter. Note that the L2 norm reduces the weight of insignificant features but does not completely remove their contribution.

\subsection{Elastic net}
The penalty-based regularization, i.e., lasso and ridge, can be further improved by increasing the number of penalty coefficients in the loss term. To this extent, Zou et al. \cite{zou_regularization_2005} elastic net regression which combines the L1 norm and L2 norm as follows:
\begin{equation}
   \hat{\theta}_{elastic}=  \operatorname*{argmin}_\theta \left\{\sum_{i=1}^{p}[y^i-(\theta_0+ \sum_{j=1}^{n} \theta_j X_j)^2] + \lambda_{1} \sum_{j=1}^{n} \lvert \theta_j \rvert + \lambda_{2} \sum_{j=1}^{n} \lvert \theta_j \rvert^{2} \right\}
\end{equation} 
Elastic net helps select the significant predictors by reducing the weights of insignificant predictors to 0.

\subsection{Random forest}
RF and XGBoost are a part of ensemble learning – a method that aggregates the results to form a multitude of decision trees. Bagging \cite{breiman_bagging_1996} and boosting \cite{bartlett_boosting_1998} are the two most widely known methods for generating trees. Boosting involves the sequential growing of trees. The successive trees utilize the information generated from the preceding trees. At each decision tree, the dataset is modified to encode the information about the errors from the previously grown trees. The successive trees give extra weight to the points that are incorrectly predicted by the preceding trees. However, in the case of bagging, each tree is grown independently — they are grown simultaneously using different bootstrapped samples of the original dataset.
For the training data, ${D_{t}}$ (Eqn \ref{training data}  ), RF works as follows: \cite{liaw_classication_2002} :\\
1. $N_{tree }$ bootstraped samples are drawn from the training data.\\ 
2. For each bootstraped sample, an unpruned decision tree is grown. Unlike a regular bagging scenario, where  $M_{random} = n $,  the RF algorithm randomly samples $M_{random}$ features, usually $\sqrt[2]{n}$ or $log_2(n)$, which are used to split each node. The best split among the sampled features is determined by information gain – gini index and entropy.
2. The final prediction( ${\hat{\psi}}_{RF}^{N_{tree}}(x)$) is made by averaging the predictions of all the decision trees as follows:
\begin{equation}
{\hat{\psi}}_{RF}^{N_{tree}}\left(x\right)=\ \frac{1}{N_{tree}}\sum_{i=1}^{N_{tree}}\psi_{RJ_{i}}(x)
\end{equation}
where ${\hat{\psi}_{RF}^{N_{tree}}\left(x\right)}$ denotes the averaged outcome of total ${N_{tree}}$ base learners and $\psi_{RF}(x)$ represents the prediction from  individual trees.\\
The error rate of RF is calculated as follows:\\
1) At each bootstrap iteration, the data points not in the bootstraped sample, called the out-of-bag (OOB) data, are predicted by the tree grown with that bootstraped sample.\\
2) OOB predictions are aggregated to estimate the OOB error rate.

\subsection{XGBoost} 
XGBoost is a boosting-based ensembling method wherein base learners (decision trees) are trained in succession to minimize the errors of the preceding trees. The training continues until a specified number of trees are grown or the objective function reaches an acceptable value. The growth of trees is governed by a greedy algorithm wherein a tree starts with 0 depth, and each tree node is split into leaves until a negative gain is reached. The following regularized objective function is minimized to learn the mapping:\\
\begin{equation}
    \Gamma=\underbrace{\sum_{i}l(\hat{y}_{i}, y_{i})}_\text{\textbf{Training loss}}  + \underbrace{\sum_{\kappa}R(f_{\kappa})}_\textbf{Regularization penalty}
    \label{Loss_func}
\end{equation}
where $l (=\frac{1}{2p}(y_{i} - \hat{y}_{i})^{2})$ is the squared loss function, $\hat{y}_{i}$ and $y_{i}$ are the predicted and the actual values for the data $\{(x_{i},\: y_{i})\}_{i=1}^{p}$. $f_{\kappa}$ represents independent tree structures. In Eqn \ref{Loss_func}, the training loss focuses on improving accuracy, and the regularization penalty tries to trade off accuracy for generalizability.
The regularization term penalizes the complexity of the tree as follows:

\begin{equation}
\underbrace{R(f_{\kappa})}_\textbf{Penalty for $k^{th}$ iteration} = \underbrace{\alpha T}_\text{Number of leaves} + \underbrace{\frac{1}{2}\beta\sum_{j=1}^{T}w_{j}^{2}}_\textbf{L2 norm of leaf scores}
\end{equation}

T is the number of leaves in a tree. $\alpha $ and $\beta$ are the hyperparameters. Using Taylor's expansion, Eqn.\ref{Loss_func} can be simplified to the following quadratic approximation for the $t^{th}$ iteration \cite{chen2016xgboost}
\begin{equation}
    \tilde{\Gamma}^{(t)} = -\frac{1}{2}\sum_{j=1}^{T}\frac{Z_{j}^{2}}{\chi_{j}+\beta} + \alpha T
    \label{simplifed objective}
\end{equation}
where, 
\begin{equation}
   Z_{j}= \sum_{i\in I_{j}}\partial_{\hat{y}_{i}^{(t-1)}}l(y_{i}, \hat{y_{i}}^{(t-1)}) 
\end{equation}

\begin{equation}
\chi_{j} = \sum_{i\in I_{j}}\partial^{2}_{\hat{y}_{i}^{(t-1)}}l(y_{i}, \hat{y_{i}}^{(t-1)})
\end{equation}
So, fundamentally, XGBoost converts the optimization of a differential objective function to determine the minimum of a quadratic equation. Each base learner is fit to the training data by optimizing Eqn. \ref{simplifed objective}.  Moreover, the regularization term in the objective equips XGBoost to prevent overfitting. The $i^{th}$ data point at $t^{th}$ iteration is predicted by adding all the previous models as follows:

\begin{equation}
    \underbrace{\hat{y}_{i}^{(t)}}_\textbf{Final model (t)} = \sum_{\kappa=1}^{t}f_{\kappa}(x_{i}) = \underbrace{\hat{y}_{i}^{(t-1)}}_\textbf{Previous model (t-1)} + \underbrace{f_{t}(x_{i})}_\textbf{New model being learnt (t)}
\end{equation}
The final output of XGBoost ($\hat{y}$) is the summation of outputs from all the K models:
\begin{equation}
    \hat{y} = \sum_{\kappa = 1}^{K}f_{\kappa}(x), f_{\kappa} \in F \text{(Functional space of base learners)}
\end{equation}

\subsection{Support vector regression (SVR)} 
SVR is an extension of the support vector machine (SVM) for regression. SVM determines a hyperplane which can linearly separate the data for classification. For data that is not linearly separable, SVM uses a mapping function, $\psi(x)$, to transform the data, $\{(x_{i}, y_{i})\}_{i=1}^{n}$, into a higher dimension. The transformed data , $\{(\phi(x_{i}), y_{i})\}_{i=1}^{n}$ , is linearly separable by the hyperplane ($f_{SVM}(x)$), expressed as\cite{guardiani2022time}:
\begin{equation}
f_{SVM}(x) = \sum_{i=1}^{n}\omega_{i}^{T}\phi_{i}(x) + b
\end{equation}
where $\omega$ is the weight vector, and b is the offset. SVM uses kernel function ($K_{i,j}$) to compute the relation between points in higher dimensions as follows:
\begin{equation}
K_{i,j} = K(x_{i}, x_{j}) = \phi(x_{i})^{T}\phi(x_{j})
\end{equation}
In this study, $K(x_{i}, x{j})$ was chosen to be radial basis function given as:
\begin{equation}
K_{i,j} = exp (-\gamma|x_{i} - x_{j}|^{2})
\end{equation}
 Unlike the aforementioned ML algorithms, which reduce the prediction error,  SVM uses a soft margin approach, which reduces the acceptable margin in which the errors can lie. SVM minimizes the following constrained regularized objective function:
\begin{equation}
    \Gamma_{SVM} = \operatorname*{min}_{\omega,b,\xi} \left(\underbrace{\frac{1}{2}\omega^{T}\omega}_\textbf{Error margin} + \underbrace{\lambda\sum_{i=1}^{n}\xi_{i}}_\textbf{Soft margin penalty} \right) \quad\text{subject to}\quad  
    \left\{
    \begin {aligned}
         & y_{i} - (\omega^{T}\phi(x_{i}) + b) \leq \epsilon + \xi_{i}, \\
         & (\omega^{T}\phi(x_{i}) + b) - y_{i} \leq \epsilon + \xi_{i},  \\
         & \xi_{i} \geq 0, \forall{i} \in \{1,...,n\}
    \end{aligned}
    \right.
    % \end{align*}
\label{Loss_func_svm}
\end{equation}
$\lambda$ is a hyperparameter. Due to soft margin formulation, slack variable $\xi$ is introduced to measure the violation of error margin for miss-classified points. The first term of the objective function minimizes the error margin, and the penalty in the second term makes the model generalizable. Lower $\lambda$ gives less importance to classification mistakes, thereby making room for relaxing the error margin. 
 \subsection{Gaussian process regression (GPR)} 
Gaussian process ($\mathcal{GP}$) is a collection of random variables (in this case, functions) such that any finite subset of the random variables has a joint Gaussian distribution. GPR is a non-parametric Bayesian ML algorithm wherein the input-output mapping function, $f\::\: X\: \rightarrow \:Y$, is assumed to be distributed as a $\mathcal{GP}$. In GPR, the output $y$ of the function $f$ for input $x$ can be written as \cite{rasmussen2006gaussian}
\begin{equation}
    y = f(x) + \epsilon_{noise}
\end{equation}
The important assumptions are that the noise term, $\epsilon_{noise}$, follows a normal distribution with $0$ mean and the function, $f(x)$, is distributed as a GP :
\begin{equation}
    \epsilon_{noise} \sim \mathcal{N} \left (0, \sigma_{noise}^{2} \right )
    \quad\text{and}\quad  
    f ( x) \sim \mathcal{GP}\,( m(x), k(x,\acute{x}))
\end{equation}
A GP is completely defined as the mean function, $m(x)$, which is the mean of all the functions in the distribution, i.e., $m(x)\: = \: \mathbb{E}[f(x)]$, and the covariance function, $k(x,\acute{x})$, which computes the correlation between function outputs at different inputs $x$ and $\acute{x}$ :
\begin{equation}
    k(x,\acute{x}) = \mathbb{E} \left[ (f(x) - m(x))(f(\acute{x}) - m(\acute{x}))\right]
\end{equation}
In this study, the Radial basis function was chosen for computing the correlation as follows:
\begin{equation}
    k_{rbf}(x,\acute{x}) = \sigma_{f}^{2}exp \left ( - \frac{\|x - \acute{x} \|^{2}}{2 \lambda^{2}} \right )
\end{equation}
where $\lambda $ and $\sigma_{f}^{2}$ are hyperparameters. If $D = \{ X_{t}, y_{t} \}$ is the training data,  the  outputs of the function, $f^{*} = [f^{*}_{i} | i = 1, ...n]^{T}$ for the new inputs $X^{*} = [x^{*}_{i} | i = 1, ...n]$, can be found from the joint normal distribution of previous observations, $y_{t}$ with the new function outputs, $f^{*}$ as follows:
\begin{equation}
    \begin{bmatrix}
        y_{t}\\
        f^{*}
    \end{bmatrix}
    \sim
    \mathcal{N} \left( 0, \begin{bmatrix}
        K(X_{t},X_{t}) + \sigma_{error}^{2}I & K(X_{t}, X^{*}) \\
        K( X^{*}, X_{t}) & K( X^{*},  X^{*})
        \label{joint_dist}
    \end{bmatrix} \right)
\end{equation}
where $K(X^*, X^*)$ is the covariance matrix between all the observed points,
\begin{equation}
    K(X^{*}, X^{*}) = \begin{bmatrix}
        k(x_{1}^{*},x_{1}^{*}) & k(x_{1}^{*},x_{2}^{*}) &\ldots & k(x_{1}^{*},x_{n}^{*})\\
        (x_{2}^{*},x_{1}^{*}) & k(x_{2}^{*},x_{2}^{*}) &\ldots & k(x_{2}^{*},x_{n}^{*})\\
        \vdots& \vdots & \ddots& \vdots\\
        (x_{n}^{*},x_{1}^{*}) & k(x_{n}^{*},x_{2}^{*}) &\ldots & k(x_{n}^{*},x_{n}^{*})\\
    \end{bmatrix}
\end{equation}
and likewise for $K(X_{t}, X^{*})$ and $K(X^{*}, X^{*})$. Using conditional distribution corresponding to Eqn\ref{joint_dist}, the predictive equations for GPR can be written as \cite{rasmussen2006gaussian}:
\begin{equation}
    f^{*} | X_{t}, y_{t}, X^{*} \sim \mathcal{N} \left( \underbrace{ m(f^{*})}_\textbf{Predective mean}, \underbrace{ k(f^{*})}_\textbf{Predective covarience}\right)
\end{equation}

\begin{equation}
\quad\text{where}\quad  m(f^{*}) = K(X^{*}, X_{t})[K(X_{t}, X_{t}) + \sigma_{noise}^{2}I]^{-1} y_{t}
\end{equation}

\begin{equation}
 \quad\text{and}\quad  k(f^{*}) = K(X^{*}, X^{*}) - K(X^{*}, X_{t})[K(X_{t}, X_{t}) + \sigma_{noise}^{2}I]^{-1} K(X_{t}, X^{*})
\end{equation}
\subsection{ Neural network (NN) } 
NN is a non-parametric mathematical model inspired by the biological functioning of neurons in the human brain.  The architecture of NN consists of an input layer, hidden layers and an output layer. The number of neurons in each layer is independent of each other. Each neuron has an activation function, $\mathbb{A}(\cdot)\: : \:\mathbb{R} \:\rightarrow \: \mathbb{R}$, which decides whether the neuron  activates on signal reception. The neurons of the subsequent layers are connected via weighted links. The neurons of the input layer receive the input signal, and the output of the input layer is propagated to the neurons of hidden layers via the links. The output, $O_{i}$, of a neuron, $i$,  in a hidden layer can be expressed as \cite{anoop_krishnan_predicting_2018}:
\begin{equation}
O_{i} = \mathbb{A} \left( \sum_{j=1}^{N}\omega_{ij}x_{j} + T_{i}^{hid} \right)
\end{equation}
where $N$ is the number of neurons in the preceding $i-1$ layer, $\omega_{ij}x_{j}$ is the weighted sum of inputs to the neuron, $i$, from the neuron of preceding layer and $T_{i}^{hid}$ is the threshold of the hidden neuron $i$ . To account for the nonlinear input-output relation, a rectified linear unit (ReLU) activation function has been used in this study and is expressed as:
\begin{equation}
\quad \mathbb{A(\text{X})} \:=\quad  
    \left\{
    \begin {aligned}
         & 0 & \mbox{for} & X < 0 , \\
         & X & \mbox{for} & X \geq 0
    \end{aligned}
    \right.
\end{equation}

\section{Hyperparametric optimization and model training} \label{appendix model training}
The optimal values for the hyper parameters obtained after performing the 4 fold cross validation is shown in Table \ref{Table_supp}
\begin{table}[h!tbp]
\setlength{\tabcolsep}{1.2pt}
\caption{\label{Table_supp} \centering{\textbf{Hyperparametric optimization}.Optimized hyperparameters for ML models predicting alite, belite, and ferrite, obtained using GridSearchCV.
}}
\centering
\resizebox{1.1\textwidth}{!}{%
\begin{tabular}{ccccc}
\hline
\multirow{2}{*}{ \textbf{Model}} &
  \multirow{2}{*}{ \textbf{Hyper parameters}} &
  \multicolumn{3}{c}{ \textbf{Optimized value}} \\
 &
   &
  \textbf{Alite} &
   \textbf{Belite} &
   \textbf{Ferrite} \\ \hline
\textbf{Lasso} &
  alpha &
  0.00001 &
  0.0001 &
  0.00001 \\ \hline
\textbf{Ridge} &
  alpha &
  0.001 &
  0.001 &
  0.00001 \\ \hline
\textbf{Elastic net} &
  alpha &
  0.00001 &
  0.00001 &
  0.00001 \\ \hline
\multirow{9}{*}{\textbf{Random forest}} &
  n\_estimators &
  300 &
  300 &
  700 \\
 &
  random\_state &
  0 &
  0 &
  0 \\
 &
  max\_depth &
  12 &
  10 &
  15 \\
 &
  n\_jobs &
  -1 &
  -1 &
  -1 \\
 &
  ccp\_alpha &
  0.0001 &
  0.0005 &
  0.0001 \\
 &
  max\_features &
  0.75 &
  0.75 &
  0.75 \\
 &
  bootstrap &
  True &
  True &
  True \\
 &
  min\_samples\_leaf &
  4 &
  6 &
  3 \\
 &
  min\_samples\_split &
  8 &
  10 &
  8 \\ \hline
\multirow{10}{*}{ \textbf{XGBoost}} &
  random\_state &
  5 &
  5 &
  5 \\
 &
  n\_estimators &
  500 &
  450 &
  600 \\
 &
  learning\_rate &
  0.03 &
  0.03 &
  0.02 \\
 &
  max\_depth &
  6 &
  6 &
  7 \\
 &
  min\_child\_weight &
  10 &
  11 &
  8 \\
 &
  subsample &
  0.8 &
  0.8 &
  0.9 \\
 &
  colsample\_bytree &
  0.8 &
  0.8 &
  0.9 \\
 &
  reg\_lambda &
  7 &
  8 &
  5 \\
 &
  reg\_alpha &
  3 &
  3 &
  1 \\
 &
  n\_jobs &
  -1 &
  -1 &
  -1 \\ \hline
\multirow{3}{*}{ \textbf{SVR}} &
  kernel &
  rbf &
  rbf &
  rbf \\
 &
  gamma &
  0.001 &
  0.001 &
  0.0001 \\
 &
  C &
  100 &
  100 &
  1000 \\ \hline
\multirow{11}{*}{ \textbf{GPR}} &
  random\_state &
  3 &
  3 &
  3 \\
 &
  normalize\_y &
  False &
  False &
  False \\
 &
  alpha &
  1e-10 &
  0.1 &
  1e-10 \\
 &
  n\_restarts\_optimizer &
  0 &
  10 &
  0 \\
 &
  kernel\_\_k1 &
  RBF(length\_scale=1) &
  $1^2$ &
  RBF(length\_scale=1) \\
 &
  kernel\_\_k2 &
  WhiteKernel(noise\_level=1) &
  RBF(length\_scale=10) &
  WhiteKernel(noise\_level=1) \\
 &
  kernel\_\_k1\_\_length\_scale &
  1 &
  1 &
  1 \\
 &
  kernel\_\_k1\_\_length\_scale\_bounds &
  (1e-5, 1e5) &
  (0.01, 1000.0) &
  (1e-5, 1e5) \\
 &
  kernel\_\_k2\_\_noise\_level &
  1 &
  10 &
  1 \\
 &
  kernel\_\_k2\_\_noise\_level\_bounds &
  (1e-5, 1e5) &
  (0.001, 1000.0) &
  (1e-5, 1e5) \\
 &
  kernel &
  \begin{tabular}[c]{@{}c@{}}RBF(length\_scale=1) +\\  WhiteKernel(noise\_level=1)\end{tabular} &
  12* RBF(length\_scale=10) &
  \begin{tabular}[c]{@{}c@{}}RBF(length\_scale=1) + \\ WhiteKernel(noise\_level=1)\end{tabular} \\ \hline
\multirow{9}{*}{ \textbf{NN}} &
  epochs &
  1000 &
  1000 &
  100 \\
 &
  batch\_size &
  256 &
  128 &
  256 \\
 &
  n\_layers &
  1 &
  1 &
  1 \\
 &
  drop &
  True &
  True &
  True \\
 &
  drate &
  0.4 &
  0.3 &
  0.3 \\
 &
  norm &
  false &
  false &
  false \\
 &
  activation &
  ReLU &
  ReLU &
  LeakyReLu \\
 &
  opt &
  Adam &
  Adam &
  SGD \\
 &
  opt\_params &
  \begin{tabular}[c]{@{}c@{}}lr : 5.815e-5 , \\ weight decay:\\ 4.811e-5\end{tabular} &
  \begin{tabular}[c]{@{}c@{}}lr:6.147e-5,\\ weight\_decay: \\ 9.940e-5\end{tabular} &
  \begin{tabular}[c]{@{}c@{}}lr:4.459e-4,\\ momentum:\\ 1.193e-1\end{tabular} \\
 &
  layers &
  116 &
  106 &
  56 \\ \hline
\end{tabular}%
}
\end{table}

The optimal values for the hyper parameters mentioned in Table \ref{Table_supp} were obtained by performing four fold cross validation on the train data set using GridSearchCV library \cite{noauthor_sklearnmodel_selectiongridsearchcv_nodate} in python. The optimized hyper parameter thus obtained were evaluated on the validation set before finally testing them on the unseen test set. Hyper parameters not mentioned in Table\ref{Table_supp} were taken as per their default values. The comparison of the model performance on the train, val and test is shown in Fig \ref{train_val_test} below.

\begin{figure}[h!tbp]
\resizebox{\textwidth}{!}{%
    \centering
    \includegraphics[scale=1.0]{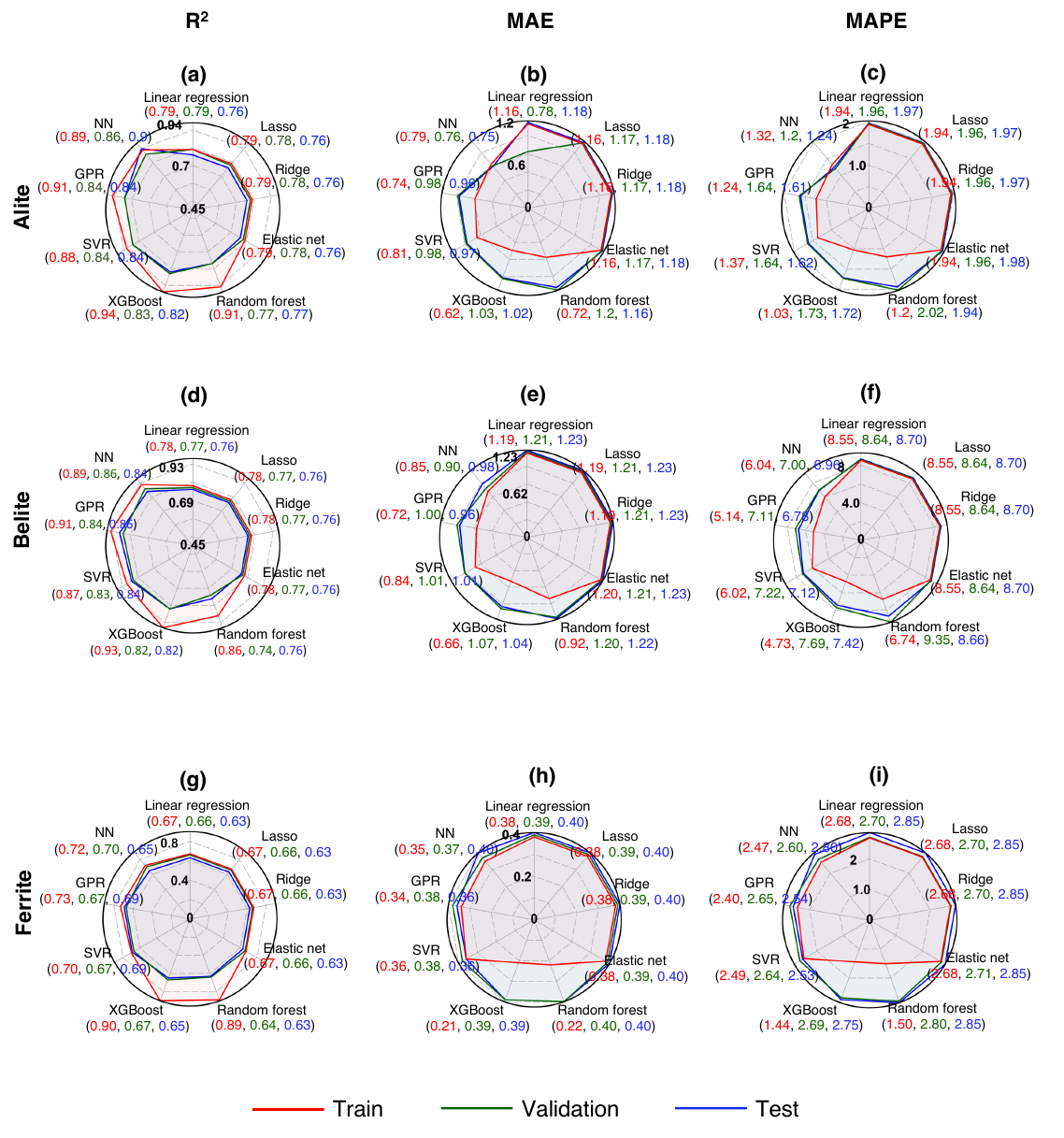}}
    \caption{\textbf{Performance comparison of machine learning architectures on training (red), validation (green), and test (blue) sets}. a–c: Alite, with metrics for $R^2$, MAE, and MAPE, respectively. d-f: Belite, with metrics for $R^2$, MAE, and MAPE, respectively. g-i: Ferrite, with metrics for $R^2$, MAE, and MAPE, respectively.}
    \label{train_val_test}
\end{figure}

\begin{figure}[h!tbp]
\resizebox{\textwidth}{!}{%
    \centering
    \includegraphics[scale=1.0]{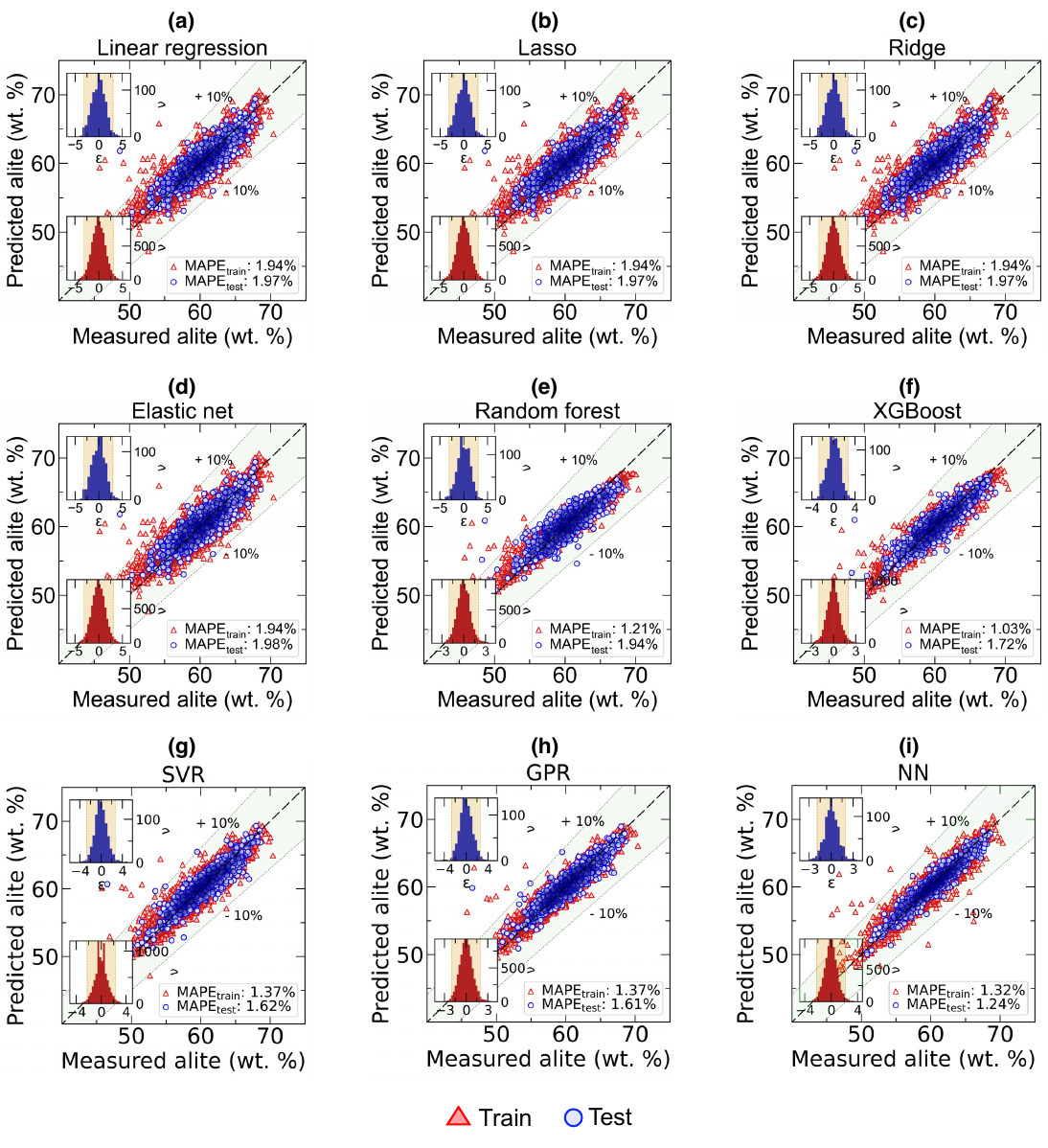}}
    \caption{\textbf{Parity plots for comparing predicted versus measured clinker phases for different ML architectures}:\textbf{a} linear regression, \textbf{b} lasso, \textbf{c} ridge, \textbf{d} elastic net, \textbf{e} random forest, \textbf{f} XGBoost, \textbf{g} SVR, \textbf{h} GPR and \textbf{i} NN for predicting alite on the train (red) and the test set (blue).The dashed black black line represents the reference prediction line while the green region represents the 10\% error margin from the reference.   }
    \label{pairty_alite}
\end{figure}

We are showing the comparison of the ML models on the train and the test set using the parity plots from Fig\ref{pairty_alite} to Fig\ref{pairty_ferrite}

\begin{figure}[h!tbp]
\resizebox{\textwidth}{!}{%
    \centering
    \includegraphics[scale=1.0]{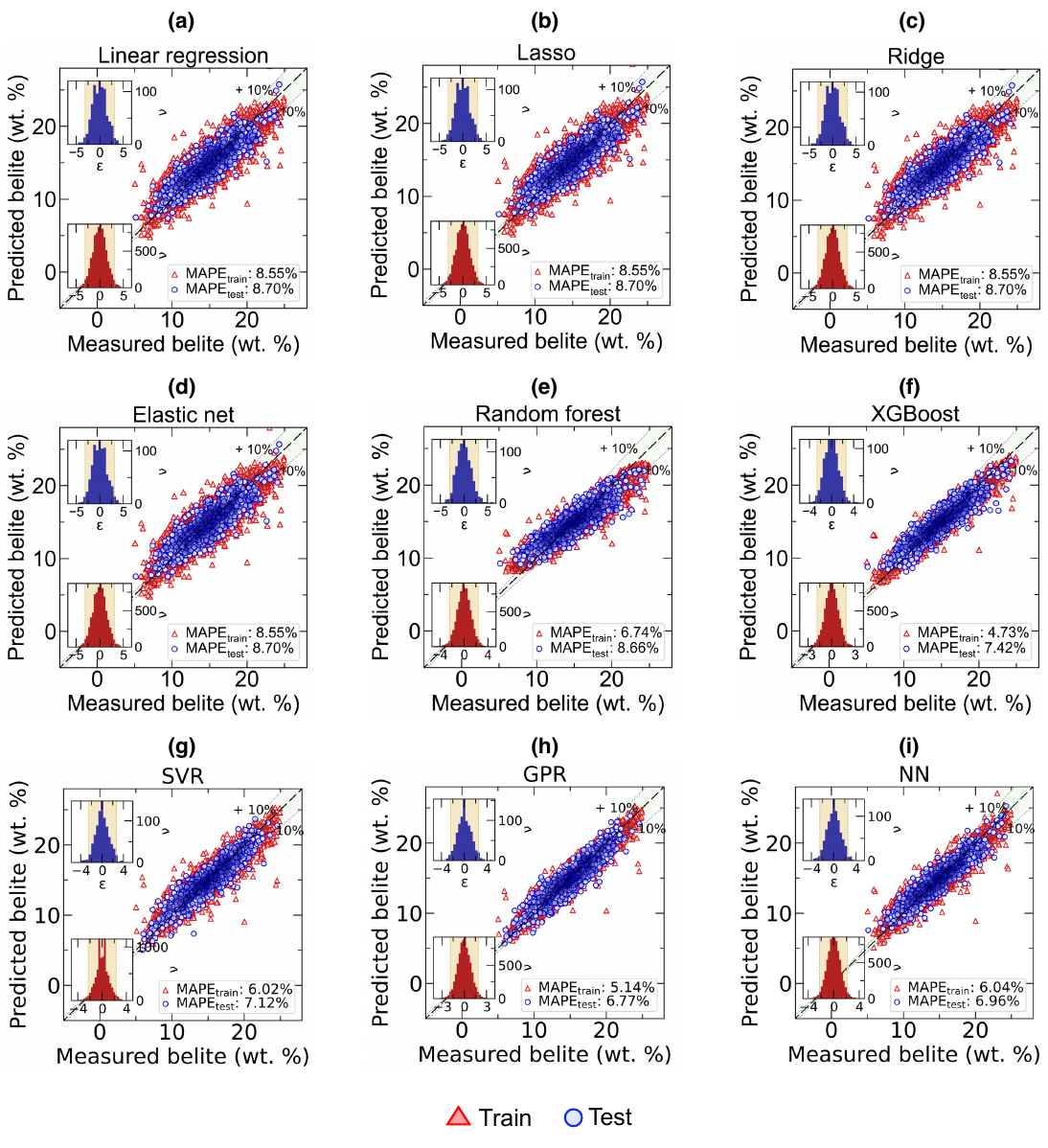}}
    \caption{Performance of \textbf{(a)} linear regression, \textbf{(b)} lasso, \textbf{(c)} ridge, \textbf{(d)} elastic net, \textbf{(e)} random forest, \textbf{(f)} XGBoost, \textbf{(g)} SVR, \textbf{(h)} GPR and \textbf{(i)} NN for predicting belite on the train and the test set.}
    \label{pairty_belite}
\end{figure}
\begin{figure}[h!tbp]
\resizebox{\textwidth}{!}{%
    \centering
    \includegraphics[scale=1.0]{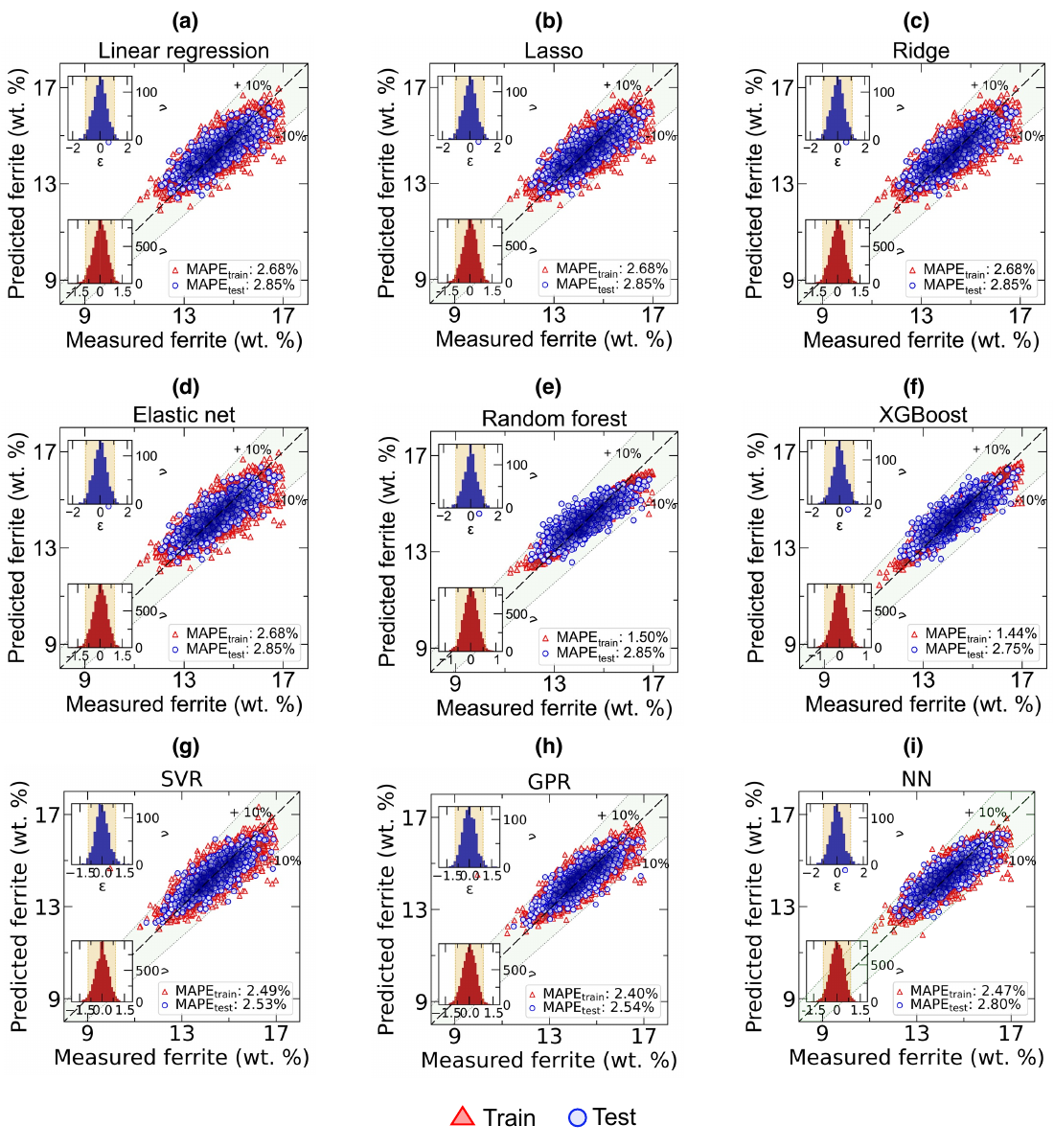}}
    \caption{Performance of \textbf{(a)} linear regression, \textbf{(b)} lasso, \textbf{(c)} ridge, \textbf{(d)} elastic net, \textbf{(e)} random forest, \textbf{(f)} XGBoost, \textbf{(g)} SVR, \textbf{(h)} GPR and \textbf{(i)} NN for predicting ferrite on the train and the test set.}
    \label{pairty_ferrite}
\end{figure}
\pagebreak

\section{Data-driven formulation of clinker equations }\label{clinker equations}
As shown in Fig. \ref{Fig2new}, the Bogue-based predictions are significantly inconsistent with the measured XRD values. Due to oversimplified assumptions used in Bogue’s approach \cite{barry_calculations_2000}, the predictions have high uncertainty in most practical situations. Though the ML models outperform Bogue, they are not as straightforward and easy to apply. Where ML models need a computational machine or at least an interface, Bogue’s prediction can be done with just pen and paper. This simplicity has kept Bogue’s approach relevant so long from $1930s$ despite the high uncertainty in its predictions. To this end, we present a data-driven approach to express the mineralogical composition of clinker as a linear algebraic equation. We used purely data-driven linear curve fitting to discover the equations from the data itself. The uncertainty in the phase estimates that is otherwise inherent to Bogue's approach due to the unrealistic assumptions is eliminated by learning the equations directly from the data. The general form of the equations formulated is as shown below:

\begin{gather}
 \underbrace{\begin{bmatrix} y_{1} \\ y_{2} \\ y_{3}\\ \end{bmatrix}} _\text{\textbf{Y}}
 =
   \underbrace{\begin{bmatrix}
   a_{11} &  \cdots &a_{19}\\
   \vdots &  \ddots  & \vdots \\
  a_{31} &  \cdots &a_{39} 
   \end{bmatrix}}_\text{\textbf{A}} 
   \underbrace{\begin{bmatrix} x_{1} \\ x_{2} \\ x_{3}\\ \end{bmatrix}}_\text{\textbf{X}}
   + \underbrace{\begin{bmatrix} b_{1} \\ b_{2} \\ b_{3}\\ \end{bmatrix}}_\text{\textbf{B}}
\end{gather}
Where $[\textbf{Y}]_{3 \times 1}$ is the clinker phase matrix, $[\textbf{A}]_{3 \times 9}$ is the coefficient matrix, $[\textbf{X}]_{3 \times 1}$ is the clinker oxides matrix, $[\textbf{B}]_{3 \times 1}$ is the intercept matrix and $y_i$, $a_{ij}$, $x_j$, $b_i$ are the elements of the matrices as indicated in equation 1. \\
The clinker equations have been developed for the following cases below.
\begin{itemize}
\item \textbf{{Case 1}}: Includes only major CO\\
In this case, the clinker phases are expressed as a linear algebraic combination of major clinker oxides. The equations developed are:
\begin{gather*}
 Alite=2.97 CaO-4.5 SiO_2-7.25 SiO_2+ 0.05Fe_2O_3 \\ 
Belite=-2.1 CaO+5.66 SiO_2+6.15 Al_2 O_3-0.11 Fe_2O_3  \\
Ferrite=0.02 CaO-0.28 SiO_2-0.28 Al_2 O_3+ 3.82Fe_2O_3   
\end{gather*}
The equations can be represented in the matrix form as follows:\\
\begin{gather*}
\begin{bmatrix} Alite \\ Belite \\ Ferrite\\ \end{bmatrix}=
 \begin{bmatrix}
   2.97 & -4.5& -7.25& 0.05 \\
   -2.1 & 5.66& 6.15& -0.11 \\
  0.02 & -0.28& -0.28& 3.82 
   \end{bmatrix}
   \begin{bmatrix} CaO \\ SiO_2 \\ SiO_2\\Fe_2O_3 \end{bmatrix}
\end{gather*}
In the subsequent cases, we gradually plug in the minor CO and the intercept into the matrix to determine how minor oxides and intercept contribute towards predicting the clinker phases. 
\item \textbf{{Case 2}}: Includes major CO and intercept.
\begin{gather*}
\begin{bmatrix} Alite \\ Belite \\ Ferrite\\ \end{bmatrix}=
 \begin{bmatrix}
   4.84 & -3.62& -4.47& 1.52 \\
  -4.71 & 4.5&2.39& -1.89 \\
  0.65 & -0.3& 0.52& 4.24 
   \end{bmatrix}
   \begin{bmatrix} CaO \\ SiO_2 \\ SiO_2\\Fe_2O_3 \end{bmatrix}+
   \begin{bmatrix}-166.9\\ -219.4 \\ -45 \end{bmatrix}
\end{gather*}

\item \textbf{{Case 3}}: Includes only major CO and minor CO.
\begin{gather*}
\begin{bmatrix} Alite \\ Belite \\ Ferrite\\ \end{bmatrix}=
 \begin{bmatrix}
3.08&-4.72&-5.59&0.09&-1.3&2.81&-13.62&4.78&-83.96\\-2.31&6.12&4.92&-0.12&-1.71&-2.83&10.76&-2.33&107\\0.11&-0.07&0.17&3.56&-0.8&2.41&-5.17&-2.85&-30
   \end{bmatrix}
   \begin{bmatrix} CaO \\ SiO_2 \\ SiO_2\\Fe_2O_3\\ MgO\\SO_3\\K_2O\\Na_2O\\Cl
   \end{bmatrix}
\end{gather*}
\item \textbf{{Case 4}}: Includes major CO, minor CO, and intercept.
\begin{align*}
\begin{bmatrix} Alite \\ Belite \\ Ferrite\\ \end{bmatrix} = &
 \begin{bmatrix}
4.72&-3.15&-4.28&3.08&0.33&2.4&-7&5.96&-75\\
-5.17&4.04&1.7&-3.48&-0.94&-5.11&3.76&-4.3&86.46\\
0.05&-0.13&0.1&3.41&-0.9&2.29&-5.35&-2.99&-27.37)
\end{bmatrix}\begin{bmatrix} CaO \\ SiO_2 \\ SiO_2\\Fe_2O_3\\ MgO\\SO_3\\K_2O\\Na_2O\\Cl\end{bmatrix} \\ &+
  \begin{bmatrix}-166.9\\ -219.4 \\ -45 \end{bmatrix}
\end{align*}
\end{itemize}

 \begin{figure}[h!tbp]

    \centering
    \includegraphics[scale=0.8]{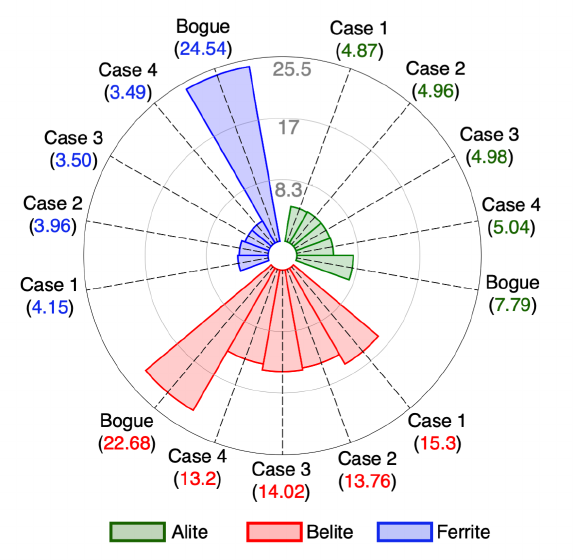}
    \caption{ Comparison of clinker equations formulated in 4 cases with Bogue’s equation for predicting clinker phases. The radial direction shows the MAPE (\%.) values}
    \label{clk equation cases}
\end{figure}

The performance of the clinker equations formulated in case $1$ to case $4$ for predicting the clinker phases is shown in Fig. \ref{clk equation cases}. It can be seen that the MAPE for case $1$ to case $4$ does not vary significantly, which indicates that plugging in additional information into the case -1 equation in the form of intercept and minor CO does not contribute significantly towards improving the prediction accuracy. However, compared to the Bogue, the developed clinker equations have remarkably fewer prediction errors. As discussed in section\ref{section ML performance}, Bogue tends to over-predict alite and under-predict belite and ferrite for the given plant. However, the developed clinker equations do not follow this specific bias. This is clear from the error histograms for the data-driven clinker equations in Fig. \ref{Fig 7} \textbf{(b)}, \textbf{(d)}and \textbf{(f)}, which are fairly symmetrical along the x-axis for all three phases.

 Altogether, the motivation behind developing the clinker equations for the plant is to develop a middle ground between Bogue’s equation-which are easy to fathom, easy to use, but high on error- and ML Models- which are not straightforward to use, infamous for working like a black box, but good at accuracy. The developed clinker equations preserve the simplicity (in terms of intelligibility and applicability) of Bogue while also making decent predictions. Though the predictive accuracy is not as remarkable as the ML models, it is still significantly better than the generic Bogue. Note that the clinker equations formulated here are plant-specific. Nonetheless, equations can be retrained for other plants also with minimal time and computation resources.

%\section{Example Appendix Section}

\setcounter{section}{0}
\renewcommand{\thesection}{\arabic{section}}
\newpage

\end{document}